%% file: neurips_2026_1.tex
\documentclass{article}

\usepackage[main, final]{neurips_2026}

\usepackage[utf8]{inputenc} 
\usepackage[T1]{fontenc}    
\usepackage{hyperref}       
\usepackage{url}            
\usepackage{booktabs}       
\usepackage{amsfonts}       
\usepackage{nicefrac}       
\usepackage{microtype}      
\usepackage{xcolor}         
\usepackage[table]{xcolor}
\usepackage{amssymb}
\usepackage{multirow}
\usepackage{adjustbox}
\usepackage{rotating}
\usepackage{graphicx}    
\usepackage{caption}     
\usepackage{cuted}       
\usepackage{enumitem}
\usepackage{makecell}
\usepackage{amsmath}
\usepackage{wrapfig} 
\usepackage{subcaption}
\usepackage[ruled]{algorithm2e}
\usepackage{afterpage}  
\title{Disentangling Dual Image References in Frequency Aware Diffusion Models for Personalized Generation}

\author{
  Haipeng Liu \quad
  Yang Wang\thanks{Yang Wang is the corresponding author} \quad
  Meng Wang \\
  School of Computer Science and Information Engineering\\Hefei University of Technology, China \\
  \texttt{hpliu\_hfut@hotmail.com}\quad
  \texttt{yangwang@hfut.edu.cn}\quad
  \texttt{eric.mengwang@gmail.com}
}

\begin{document}

\maketitle


\begin{figure*}[h]
    \centering
    \includegraphics[width=\textwidth]{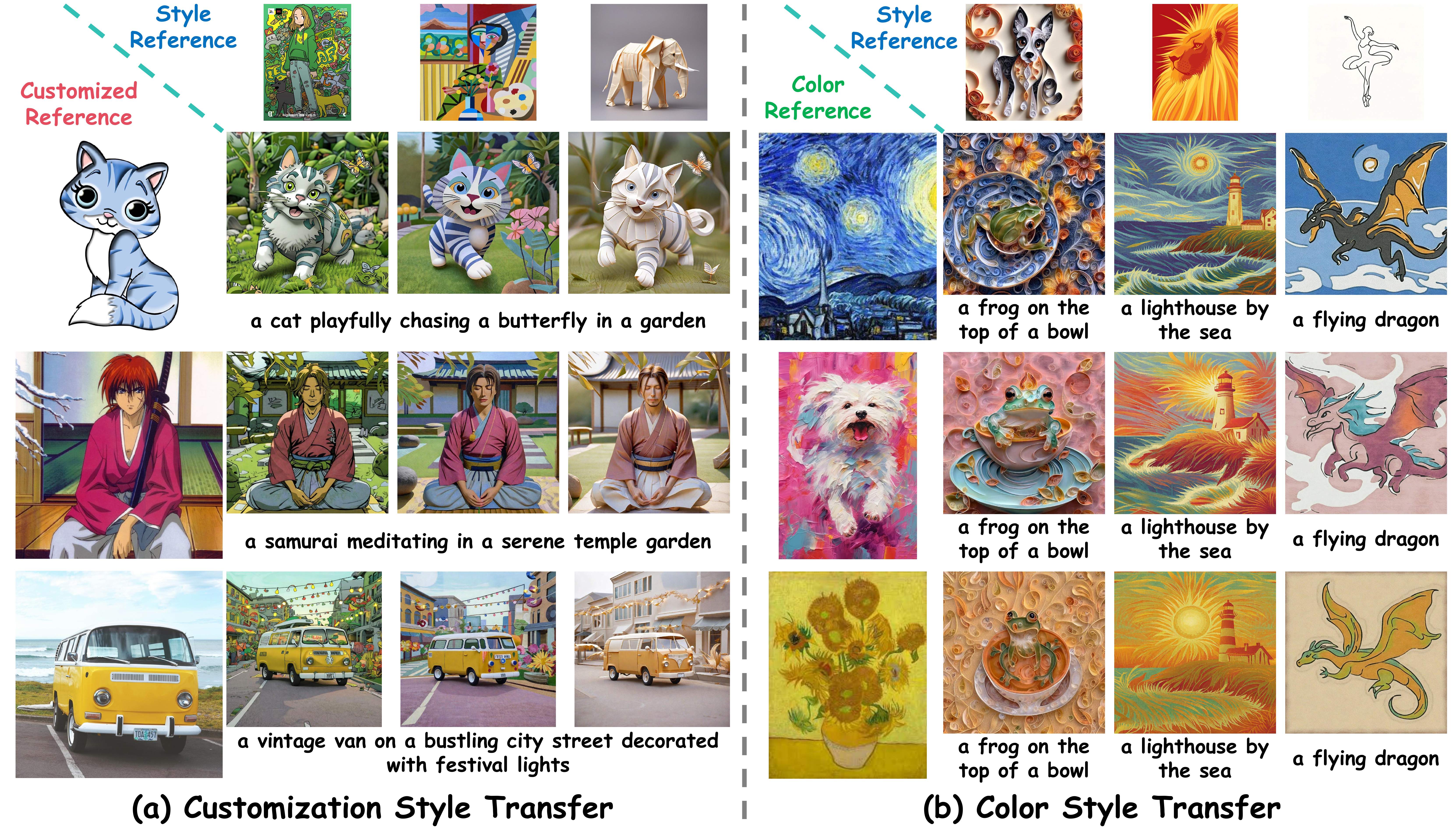}
    \caption{Personalized images generated by our \textbf{Dual-FDM}. (a) The generated examples of \textbf{Dual-FDM$_{cust}$} for customized references (left) and style references (above); (b) The generated examples of \textbf{Dual-FDM$_{cost}$} for  color references (left) and style references (above).}
    \label{fig:vis}
\end{figure*}

\begin{abstract}
Personalized image generation aims to synthesize text-driven images conditioned on reference images, while mainly casting the generation as image customization for foreground and style transfer for background. Previous arts of diffusion models suffers from the text misalignment with background for image customization and foreground for style transfer during the denoising process. Such facts, as we observed, rooted from the entanglement among hybrid frequency bands during the denoising process. In particular, the low-frequency band primarily captures color information, mid-frequency band encodes global structure, while high-frequency band corresponds to local texture.
Besides, the early (high-noise) denoising stage mainly focuses on the low-and-mid frequency bands, while the late (low-noise) denoising stage progressively exhibits the high-frequency band, which, all together, overwhelm the text prompt guidance, suffering from the heavy reliance on reference image. To address such salient limitation, in this paper, we study personalized generation based on dual references - customization and color style reference - to propose a paradigm to disentangle these \textbf{\underline{Dual}} image references within \textbf{\underline{F}}requency-aware \textbf{\underline{D}}iffusion \textbf{\underline{M}}odels, dubbed \textbf{Dual-FDM}, to simultaneously tackle two crucial personalized image generation tasks: customization style transfer and color style transfer, by disentangling different frequency bands via mask strategy within frequency domain. For customization style transfer,  we replace the mid-frequency band of the background in the style reference with that from the foreground of the customized reference. For color style transfer, we substitute the low-frequency band of the background in the style reference with that from both the foreground and background of the color reference. Both the substituted frequency bands are used as the key and value to reconstruct the query foreground and background of the denoised personalized image.
We further compute the ratio of information entropy of the substituted frequency bands to adaptively modulate the denoising timesteps between the early and late stages. Extensive experiments validate the superiority of \textbf{Dual-FDM} over the state-of-the-art diffusion models for  personalized image generation. Our code can be accessed from \textup{\url{https://github.com/htyjers/Dual-FDM}}.
\end{abstract}

\input{sec/1_introduction}

\input{sec/3_method}
\input{sec/4_experi}

\section{Conclusion}
In this paper, we  study personalized generation based on dual references - customization and color and style reference - and propose a paradigm  to disentangle dual image references within frequency-aware diffusion models, dubbed \textbf{Dual-FDM}, for two classical personalized image generation tasks in a row: customization style transfer and color style transfer, by disentangling different frequency bands via mask strategy within frequency domain.  Extensive experiments validate
the superiority of \textbf{Dual-FDM} over the state-of-the-art diffusion models.

\textbf{Acknowledgments} This research is sponsored by CCF-NetEase ThunderFire Innovation Research Funding (NO.
CCF-Netease 202513) and supported by Institute of Advanced Medicine and Frontier Technology
(2023IHM01080); The computation is completed on the HPC Platform of Hefei University of
Technology.

{
    \small
    \bibliographystyle{plain}
    \bibliography{main}
}


\newpage
\appendix

\input{sec/5_appendix}


\afterpage{\clearpage} 

\newpage
\input{checklist.tex}

\end{document}

%% file: sec/1_introduction.tex
\section{Introduction}\label{intord}
\begin{wrapfigure}{r}{0.55\textwidth} 
  \centering
  \includegraphics[width=\linewidth]{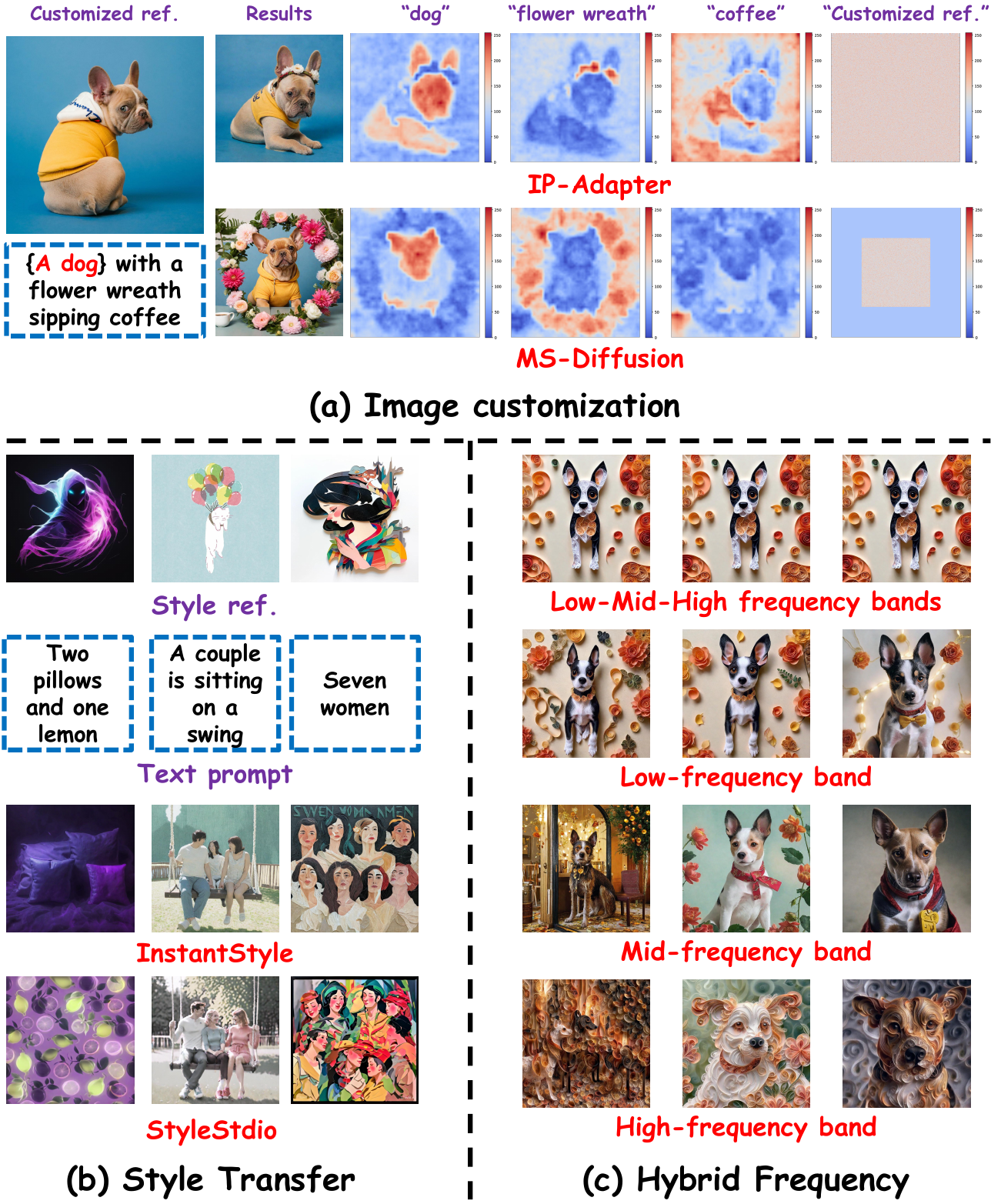}
  \caption{\small (a) Comparison of the generated results, as well as the attention score map of text prompts and image references, between IP-Adapter \cite{ye2023ip} and MS-Diffusion \cite{wangms} for image customization. (b) Comparison of the generated results between InstantStyle \cite{wang2024instantstyle} and StyleStudio\cite{lei2025stylestudio} for style transfer. (c) Comparison of generated results guided by different frequency band of the image reference.}
  \label{fig:mov}
\end{wrapfigure}

Personalized image generation focuses on synthesizing text-driven images conditioned on reference images, which enjoys remarkable progress, benefiting from the powerful denosing capabilities of Denoising Diffusion Probabilistic Models (DDPMs) \cite{ho2020denoising, liu2024structure,liu2026one}, \textit{e.g.}, Stable Diffusion \cite{9878449, esser2024scaling, podell2023sdxl,qian2022switchable,qian2023adaptive,wang2024unpacking,song2026stylevideo}. Typically, the personalized image generation involves two key subtasks, \textit{i.e.}, \textit{image customization} and \textit{style transfer}. Given text prompts, image customization \cite{xiao2025fastcomposer, zeng2024jedi, pang2024attndreambooth, gal2022image, ruiz2023dreambooth, kumari2023multi, liu2025tfcustom, ma2024subject, li2024tuning, li2023blip,liu2022delving,11626075,jiang2026rare,jiang2025cmsl} aims to generate images by substituting the foreground semantics in the text prompts with the foreground extracted from the reference images, while preserving the background of the text prompts. In contrast, style transfer \cite{qin2025freelunchcolortexturedisentanglementstylized, rout2025rb, liu2023stylecrafter, sohn2023styledrop, zhuless, qi2024deadiff, hertz2024style, lei2025stylestudio, gao2024styleshot, xing2024csgo, wang2024instantstyle1, wang2024instantstyle, jeong2024training,wang2022progressive} focuses on disentangling style (\textit{background}) from reference images and aligning the foreground with the text prompt. However, prior arts have struggled to align the generated image with the text prompts within each individual subtask.

\begin{figure}[h]
  \centering
  \includegraphics[width=\textwidth]{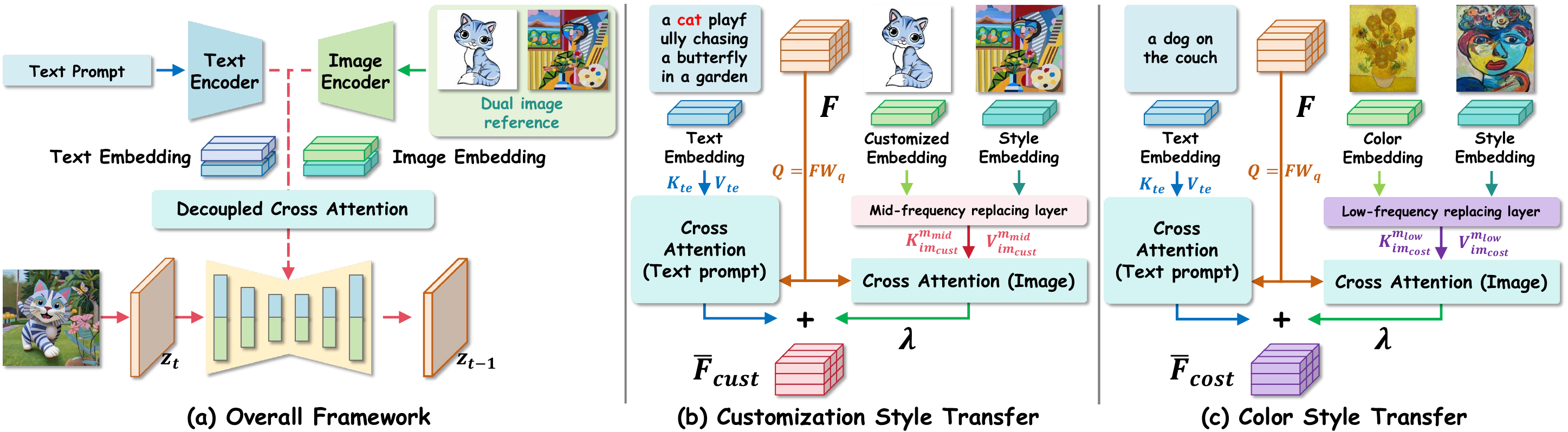}
  \caption{Illustration of our proposed \textbf{Dual-FDM} pipeline (a). For customization style transfer (b),  we replace the mid-frequency band of the background in the style reference with that from the foreground of the customized reference, named \textbf{Dual-FDM$_{cust}$} (Sec.~\ref{sec:icst}). For the color style transfer (c), we substitute the low-frequency band of the background in the style reference with that from both the foreground and background of the color reference, named \textbf{Dual-FDM$_{cost}$} (Sec.~\ref{sec:ctst}). Each of the two personalized image generation tasks are conducted in the early and late stage of the whole denoising process guided by text prompts as illustrated in Eq.~(\ref{eq:attention1}).}
  \label{fig:model}
\end{figure}

For image customization, substantial efforts \cite{ye2023ip, aiello2025dreamcache, huang2025resolving} aim to \textit{emphasize dense image references of the foreground while overwhelming sparse text prompts of the background, especially for complex ones}, \textit{e.g.}, a degraded ``\textit{flower wreath}'' and the missing ``\textit{coffee}'' in the IP-Adapter \cite{ye2023ip} of Fig.~\ref{fig:mov}(a). Building on this, layout-based methods \cite{wangms, huang2025resolving, zhang2024ssr} exploit spatial layouts to explicitly constrain reference concepts to the foreground, while assigning background regions to text prompts. \textit{This preserves the foreground identity, but the explicit separation of foreground and background in the layout disrupts overall consistency, leading to the misalignment with  (complex / long) text prompts.} As illustrated in the MS-Diffusion \cite{wangms} of Fig.~\ref{fig:mov}(a),  the generated spatial arrangement of ``\textit{flower wreath}'' and ``\textit{coffee}'' diverges from the prompt and instead resembles \textit{``A flower wreath around a dog near a coffee''}.

When coming to style transfer, during the early stage of the denoising process with high-level noise, the foreground is reconstructed from sparse text prompts, \textit{the dense style information intended for the background often leaks into the foreground, resulting in misalignment with the text prompts}, \textit{e.g.}, the incorrect number of ``\textit{woman}" in the InstantStyle \cite{wang2024instantstyle} of Fig.~\ref{fig:mov}(b). To resolve such issue, StyleStudio \cite{lei2025stylestudio} applies style via instance normalization \cite{huang2017arbitrary} in the cross-attention layers and leverages layout to preserve foreground in the self-attention layers. Nevertheless, self-attention score maps highlight only the most salient regions, implying that the extracted layout can align only with simple text prompts (\textit{i.e.}, ``\textit{a dog}")  rather than the multi-objects (``\textit{two pillows and one lemon}" in the StyleStudio \cite{lei2025stylestudio} of Fig.~\ref{fig:mov}(b)) text prompts, thereby \textit{preventing it from handling the complex text prompts}.


So far, neither image customization nor style transfer achieves satisfactory alignment between the generated image and the text prompt for diffusion models. Such facts, as we observed, rooted from the entanglement among hybrid frequency bands. Intuitively, as shown in Fig.~\ref{fig:mov}(c), the low-frequency and mid-frequency bands capture color and global structure, while high-frequency band focuses on local texture, where different frequency bands are mainly denoised during \textit{different} stages of the denoising process. As we observed, the early (high-noise) denoising stage mainly focuses on the low-and-mid frequency bands, while the late (low-noise) denoising stage progressively exhibits the high-frequency band (See Sec.~\ref{sec1s} of the Appendix), suffering from the heavy reliance on reference image rather than text prompts. 

To resolve the text misalignment for personalization generation, we propose to solve the problem upon dual references and develop a paradigm to disentangle these \textbf{\underline{Dual}} image references within \textbf{\underline{F}}requency-aware \textbf{\underline{D}}iffusion \textbf{\underline{M}}odels, dubbed \textbf{Dual-FDM}, for two classical personalized image generation tasks in a row: customization style transfer and color style transfer, by disentangling different frequency bands via mask strategy within frequency domain. Specifically:
\begin{itemize}[leftmargin=*]
\item We propose to map the foreground or background of image references into the frequency domain via Discrete Cosine Transform (DCT), where a mask is applied from  left to  right to progressively disentangle the low-, mid-, and high-frequency bands. Notably, for customization style transfer,  we replace the mid-frequency band of the background in the style reference with that from the foreground of the customized reference, named \textbf{Dual-FDM$_{cust}$} (Fig.~\ref{fig:vis}(a)). The style reference with the mid-frequency band substituted is used as the key and value to reconstruct the query foreground and background of the denoised personalized image in the decoupled cross-attention. 
    \item For the color style transfer, we substitute the low-frequency band of the background in the style reference with that from both the foreground and background of the color reference, named \textbf{Dual-FDM$_{cost}$} (Fig.~\ref{fig:vis}(b)). The style reference with the substituted low-frequency band is further utilized as the key and value to reconstruct the query foreground and background of the denoised personalized image in the decoupled cross-attention layer.
\item  To pop up the foreground or background of the image reference, we utilize the key projection matrices  in the decoupled cross-attention layer to represent the foreground, which are then reformulated through singular value decomposition (SVD) to derive the corresponding \textit{background matrix} for the background.
\item  For each task, we further compute the ratio of information entropy of the substituted frequency bands to adaptively modulate the denoising timesteps between the early and late stages, so as to achieve a well-balanced performance between text alignment and dual-image similarity.
\end{itemize}
Extensive experiments validate the superiority of Dual-FDM over the state-of-the-art diffusion models for  personalized image generation.

%% file: sec/3_method.tex
\section{Methodology}\label{method}
Central to our proposed Dual-FDM cover the following aspects: 1) how to pop up  the background of the image references (Sec.~\ref{sec:back}); 2) how to disentangle the mid-frequency band in customization style transfer (Sec.~\ref{sec:icst}); 3) how to disentangle the low-frequency band in color style transfer (Sec.~\ref{sec:ctst}); 4) how to adaptively modulate the timesteps within the denoising process between early and late stages (Sec.~\ref{sec:ada}). Before shedding light on Dual-FDM, we elaborate the preliminary regarding decoupled cross-attention mechanism, which plays a pivotal role of personalized image generation. 
\subsection{Preliminary}
Often, personalized image generation methods adopt a decoupled cross-attention mechanism (\textit{i.e.}, \textbf{IP-Adapter}) to integrate text prompt $C_{te} \in \mathbb{R}^{s_{te} \times d_{te}}$, the reference image $C_{im} \in \mathbb{R}^{s_{im} \times d_{im}}$ into every cross-attention layer of the denoising network where $s_{te}$ and $s_{im}$ denote the token number of the text prompts and image patches, such that $d_{te}$ and $d_{im}$ are the dimension for token and patch representation.  The denoised latent image feature $F \in \mathbb{R}^{(h \times w) \times d}$ serves as the query to be reconstructed by the key of the text prompts via the $l$-th cross-attention layers for the UNet architecture $\epsilon_{\theta}(F, t, C_{te}, C_{im})$, as formulated below:
\begin{equation}
\label{eq:attention1}
\begin{split}
\overline{F}_{te} = Atten(Q, K_{te}, V_{te}) = \text{Softmax}\left(\frac{QK_{te}^T}{\sqrt{d_{te}}}\right)V_{te},\\
\end{split}
\end{equation}
where $Q \in \mathbb{R}^{(h \times w) \times d}$ denotes the query matrix, such that $Q = F\text{W}_q$ with $\text{W}_q \in \mathbb{R}^{d \times d}$ being as trainable linear projection matrix. Likewise, $K_{te} \in \mathbb{R}^{s_{te} \times d}$ and $V_{te} \in \mathbb{R}^{s_{te} \times d}$ denote the key and value matrices, such that $K = C_{te}\text{W}_k^{te}$ and $V = C_{te}\text{W}_v^{te}$ with $\text{W}_k^{te} \in \mathbb{R}^{d_{te} \times d}$ and $\text{W}_v^{te} \in \mathbb{R}^{d_{te} \times d}$ being as trainable linear projection matrices. Meanwhile, $F$ is fed into another cross-attention module to interact with the reference image embeddings below: 
\begin{equation}
\label{eq:attention2}
\begin{split}
\overline{F}_{im} = Atten(Q, K_{im}, V_{im}) = \text{Softmax}\left(\frac{QK_{im}^T}{\sqrt{d_{im}}}\right)V_{im},\\
\end{split}
\end{equation}
where $K_{im} \in \mathbb{R}^{s_{im} \times d}$ and $V_{im} \in \mathbb{R}^{s_{im} \times d}$ denote the key and value matrices, such that $K = C_{im}\text{W}_k^{im}$ and $V = C_{im}\text{W}_v^{im}$ with $\text{W}_k^{im} \in \mathbb{R}^{d_{im} \times d}$ and $\text{W}_v^{im} \in \mathbb{R}^{d_{im} \times d}$ being as trainable linear projection matrices. The predicted noise, \textit{i.e.}, output of $\epsilon_{\theta}(F, t, C_{te}, C_{im})$, is then subtracted from $F$ to update the latent feature, such denoising process iterates until $t=0$, yielding the final  image.


%

\subsection{Popping up  the Background of Image Reference}\label{sec:back}
Given text prompts, the major bottleneck for disentangling the dual image references lies in mapping the foreground or background of image references into the frequency domain. To this end, our basic idea is to pop up the background of the image reference, which paves the way to achieve the text alignment with two classical personalized image generation tasks in a row.

\begin{figure}[t]
\centering
\begin{minipage}{0.59\linewidth}
  \centering
  \includegraphics[width=\linewidth]{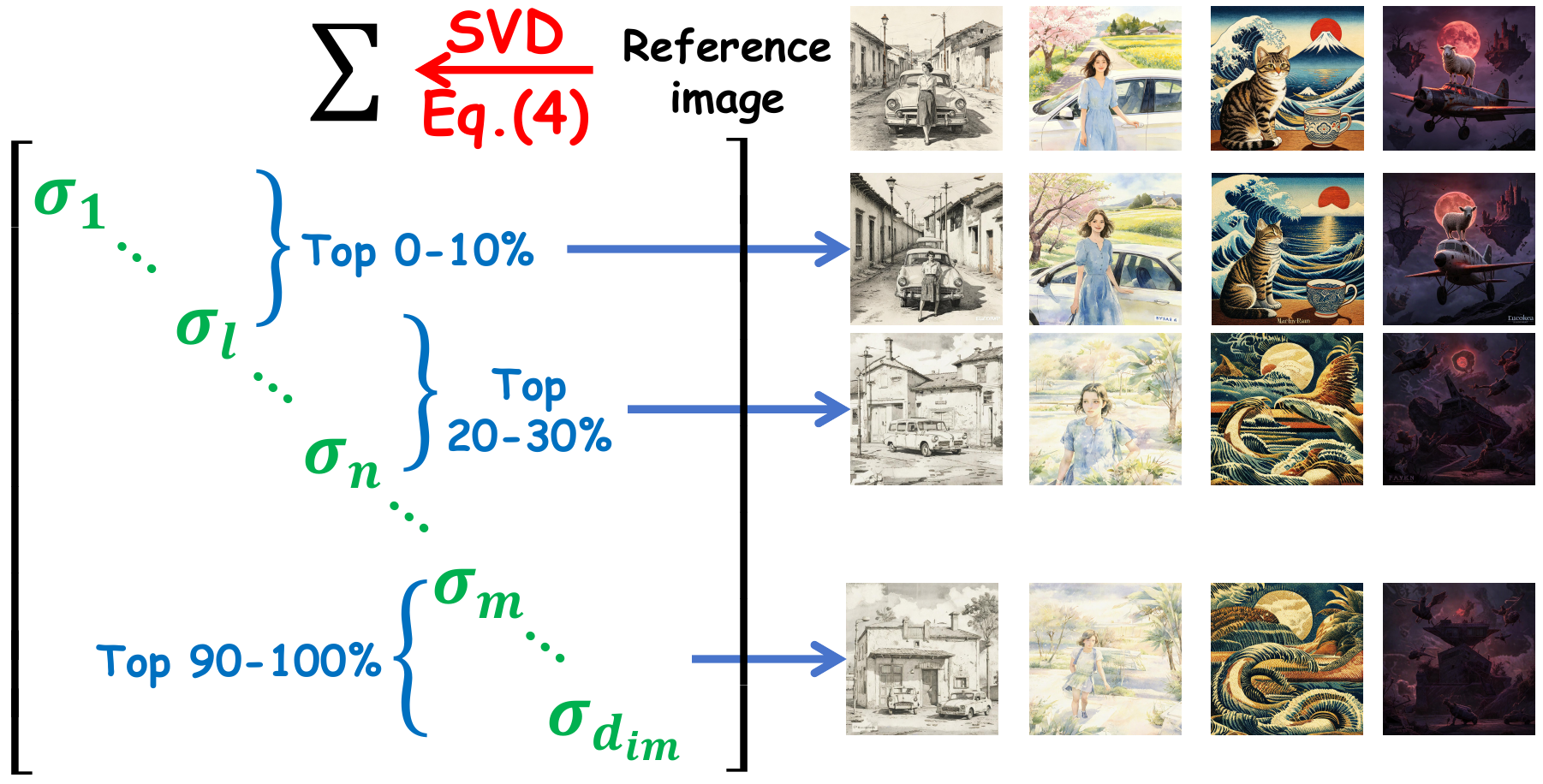}
  \caption{We visualize image components reconstructed from different singular value intervals of $W_k^{im}$. The reference image from first two columns correspond to 3D spatial relationships ("\textit{in front of}"), and the last two to 2D cases ("\textit{on right of}" and "\textit{on top of}"). Larger singular values (0–10\%) mainly capture foreground, while smaller ones (90–100\%) correspond to background. In 3D scenes, varying saliency leads singular values in 20–30\% to represent partial objects. In contrast, 2D scenes exhibit more uniform saliency across targets.}
  \label{fig:mov0}
\end{minipage}
\hfill
\begin{minipage}{0.39\linewidth}
  \centering
  \includegraphics[width=\linewidth]{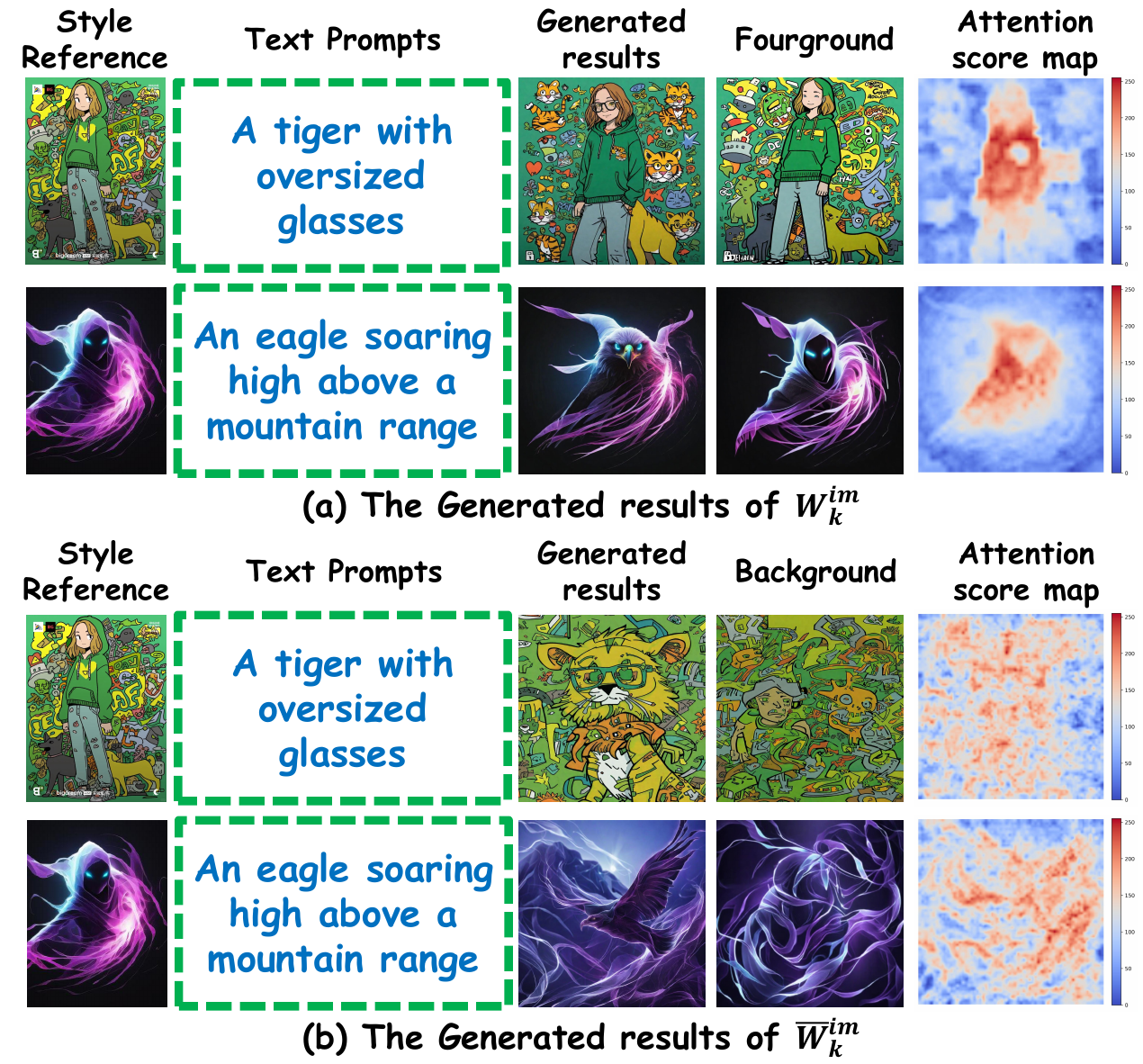}
  \caption{The style transfer generated results using (a) $W_k^{im}$ (Eq.~(\ref{eq:cor})), and (b) background matrix $\overline{W}_k^{im}$ (Eq.~(\ref{eq:style})), which demonstrates that the background from the image reference predominantly contributes to reconstructing the denoised image.}
  \label{fig:mov1}
\end{minipage}
\end{figure}

Following Eq.~(\ref{eq:attention2}), to obtain the background of the image reference within the decoupled cross-attention layer, we visualize the generated result of $\text{W}_k^{im}$ to analyze the regions where its contributions are predominantly concentrated, as illustrated in the last column of Fig.~\ref{fig:mov1}(a), where $\text{W}_k^{im}$ indicates that the attention scores among different patches effectively encode their foreground. To mitigate such issue, we perform singular value decomposition (SVD) on $\text{W}_k^{im}$:
\begin{equation}
\label{eq:cor}
\begin{aligned}
&W_k^{im} = U \Sigma V^T, U \in \mathbb{R}^{d_{im} \times d_{im}}, V\in \mathbb{R}^{d \times d_{im}}, \Sigma = \operatorname{diag}\!\left(\sigma_1, \sigma_2, \dots, \sigma_{d_{im}}\right) \in \mathbb{R}^{d_{im} \times d_{im}},
\end{aligned}
\end{equation}
where all singular values $\sigma_i$ in $\Sigma$ indicate the relative importance of each component. We leverage distinct singular components to disentangle the background and foreground from the  reference image, following the observations in \textbf{Remark 1}.
%
%
%

\noindent \textbf{Remark 1}. \textbf{Larger $\sigma_i$ tend to correspond to the foreground, whereas smaller $\sigma_i$ are more likely associated with the background}. 

As illustrated in Fig.~\ref{fig:mov0}, SVD decomposes the image feature into components corresponding to different saliency levels, where components associated with larger singular values represent more salient structures in the image. \hfill $\blacksquare$

To suppress the foreground components and enhance the background components, we invert the singular values and construct the background matrix $\overline{W}_k^{im}$ as follows:
\begin{equation}
\label{eq:style}
\begin{aligned}
&\overline{W}_k^{im} = U \operatorname{diag}\!\left(\tfrac{1}{\sigma_1}, \tfrac{1}{\sigma_2}, \dots, \tfrac{1}{\sigma_{d_{im}}}\right) V^T,\quad \sigma_i \neq 0.
\end{aligned}
\end{equation}
We remark the intuition of Eq.~(\ref{eq:style}) as follows:

\noindent \textbf{Remark 2}. \textbf{Inverting the singular values suppresses dominant foreground components while enhancing background-related information}. 

Eq.~(\ref{eq:style}) provides an intuitive means of disentangling the background from the foreground in the cross-attention layer. As illustrated in the fourth column of Fig.~\ref{fig:mov1}(b), $\overline{W}_k^{im}$ suppresses the foreground and retains background-related information. \textit{\textbf{Due to page limitation, we provide more intuitions for the rationale between ${W}_k^{im}$ and $\overline{W}_k^{im}$ in Sec.~\ref{sec2} of the Appendix.}}  \hfill $\blacksquare$


\subsection{Customization Style Transfer: Replacing the Mid-frequency of Image Reference}\label{sec:icst}
\begin{figure}[t]
  \centering
  \includegraphics[width=\linewidth]{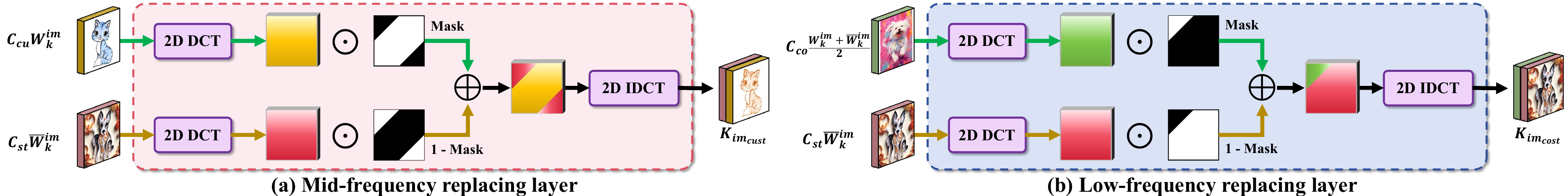}
  \caption{Illustration of  (a) mid-frequency replacing layer and (b) low-frequency replacing layer.}
  \label{fig:module}
\end{figure}

Following  $W_k^{im}$ (Eq.~(\ref{eq:cor})) and $\overline{W}_k^{im}$ (Eq.~(\ref{eq:style})), we initialize the denoising process with random Gaussian noise $z_{T}$, together with the text prompts $C_{te}$ to describe both foreground and background, the customized reference $C_{cu} \in \mathbb{R}^{s_{im} \times d_{im}}$, and the style reference $C_{st} \in \mathbb{R}^{s_{im} \times d_{im}}$, as inputs to the denoising network $\epsilon_{\theta}(F, t, C_{te}, C_{cu}, C_{st})$ to generate the denoised results for both the image customization and style transfer, then we reformulate Eq.~(\ref{eq:attention2}) as follows:
\begin{equation}
\label{eq:attention4}
\begin{split}
\overline{F}_{im_{cust}} &= Atten(Q, K^{m_{{mid}}}_{im_{cust}}, V^{m_{{mid}}}_{im_{cust}}) = \text{Softmax}\left(\frac{Q(K^{m_{{mid}}}_{im_{cust}})^T}{\sqrt{d_{im}}}\right)V^{m_{{mid}}}_{im_{cust}},\\
\end{split}
\end{equation}
where  $K^{m_{{mid}}}_{im_{cust}} \in \mathbb{R}^{s_{im} \times d}$ and $V^{m_{{mid}}}_{im_{cust}} \in \mathbb{R}^{s_{im} \times d}$  denote the key and value matrices obtained by replacing the mid-frequency bands of the background in the style reference with those from the foreground of the customized reference via mid-frequency replacing layer (see Fig.~\ref{fig:module}(a)), as formulated below:
\begin{equation}
\label{eq:mfbs}
\begin{aligned}
K^{m_{{mid}}}_{im_{cust}} &= \text{IDCT}\Bigl(
\text{DCT}(C_{cu} W_k^{im}) \odot m_{{mid}} +
\text{DCT}(C_{st} \overline{W}_k^{im}) \odot (1 - m_{{mid}})
\Bigr), \\
V^{m_{{mid}}}_{im_{cust}} &= \text{IDCT}\Bigl(
\text{DCT}(C_{cu} W_v^{im}) \odot m_{{mid}}+
\text{DCT}(C_{st} W_v^{im}) \odot (1 - m_{{mid}})
\Bigr),
\end{aligned}
\end{equation}
where DCT($\cdot$) represents Discrete Cosine Transform, IDCT($\cdot$) represents Inverse DCT, $ \odot $ is the element-wise multiplication and $ m_{{mid}} $ denotes the mid-pass mask to isolate mid-frequency band, the denoised mid-frequency band layer employs a binary mask defined by the sum of the 2D coordinates as a threshold below:
\begin{equation}
m_{{mid}}(x,y) =
\begin{cases}
1, & \text{if } th_{mp1} < x + y \leq th_{mp2} \\
0, & \text{otherwise}
\end{cases}
\label{eq:mid1}
\end{equation}
where $ th_{mp1} $ and $ th_{mp2} $ are the thresholds for mid-pass filtering to extract mid-frequency band.  Following the above intuition, the final output of the decoupled cross-attention is  $\overline{F}_{{cust}} = \overline{F}_{te} + \lambda\overline{F}_{im_{cust}}$, such that $\lambda \in [0,1]$ is a balancing weight. 

\subsection{Color Style Transfer: Replacing the Low-frequency of Image Reference}\label{sec:ctst}
Based on  $W_k^{im}$ (Eq.~(\ref{eq:cor})) and $\overline{W}_k^{im}$ (Eq.~(\ref{eq:style})), we initialize the denoising process with random Gaussian noise $z_{T}$, together with the text prompts $C_{te}$ to describe both  foreground and background, the color reference $C_{co} \in \mathbb{R}^{s_{im} \times d_{im}}$, and the style reference $C_{st} \in \mathbb{R}^{s_{im} \times d_{im}}$, as inputs to the denoising network $\epsilon_{\theta}(F, t, C_{te}, C_{co}, C_{st})$ to generate the denoised results for both the color transfer and style transfer, then we reformulate Eq.~(\ref{eq:attention2}) as follows:
\begin{equation}
\label{eq:attention4}
\begin{split}
\overline{F}_{im_{cost}} &= Atten(Q, K^{m_{{low}}}_{im_{cost}}, V^{m_{{low}}}_{im_{cost}}) = \text{Softmax}\left(\frac{Q(K^{m_{{low}}}_{im_{cost}})^T}{\sqrt{d_{im}}}\right)V^{m_{{low}}}_{im_{cost}},\\
\end{split}
\end{equation}
where  $K^{m_{{low}}}_{im_{cost}} \in \mathbb{R}^{s_{im} \times d}$ and $V^{m_{{low}}}_{im_{cost}} \in \mathbb{R}^{s_{im} \times d}$  denote the key and value matrices obtained by replacing the low-frequency bands of the background in the style reference with those from the foreground and the background of the color reference via low-frequency replacing layer (see Fig.~\ref{fig:module}(b)), and is then formulated as:
\begin{equation}
\label{eq:lfbs}
\begin{aligned}
K^{m_{{low}}}_{im_{cost}} &= \text{IDCT}\Bigl(
\text{DCT}\!\left(C_{co}\!\left(\tfrac{W_k^{im} + \overline{W}_k^{im}}{2}\right)\right) \odot m_{{low}}  +
\text{DCT}(C_{st} \overline{W}_k^{im}) \odot (1 - m_{{low}})
\Bigr), \\
V^{m_{{low}}}_{im_{cost}} &= \text{IDCT}\Bigl(
\text{DCT}(C_{co} W_v^{im}) \odot m_{{low}} +
\text{DCT}(C_{st} W_v^{im}) \odot (1 - m_{{low}})
\Bigr),
\end{aligned}
\end{equation}
where $ m_{{low}} $ is defined as the low-pass mask below:
\begin{equation}
\label{eq:mask1}
m_{{low}}(x,y) =
\begin{cases}
1, & \text{if } x + y \leq th_{lp} \\
0, & \text{otherwise}
\end{cases}
\end{equation}
where $ th_{lp} $ is the threshold for low-pass filtering.  Following the above, the final output  of the decoupled cross-attention is $\overline{F}_{{cost}} = \overline{F}_{te} + \lambda\overline{F}_{im_{cost}}$.

\subsection{Adaptively Modulating the Early and Late Denoising Stage}\label{sec:ada}
To adaptively \textit{modulate} the early and late stages, we compute the threshold $\Delta_{cust}$ for customization style transfer and $\Delta_{cost}$ for color style transfer as follows:
\begin{equation}
\begin{aligned}
\label{eq:ratio}
&\Delta_{cust} =
\frac{H\!\left(K^{m^{e}_{mid}}_{im_{cust}}\right)}
     {H\!\left(K^{m^{l}_{mid}}_{im_{cust}}\right)}, \quad
\Delta_{cost} =
\frac{H\!\left(K^{m^{e}_{low}}_{im_{cost}}\right)}
     {H\!\left(K^{m^{l}_{low}}_{im_{cost}}\right)}, \\
& H(x) = - \sum_{i=1}^{d}
\operatorname{Softmax}(x)_i
\log\!\left( \operatorname{Softmax}(x)_i + \epsilon \right),
\end{aligned}
\end{equation}
where $H(\cdot)$ denotes the information entropy of the mixed image references, and $\epsilon$ is a small constant to ensure numerical stability.
$m^{e}_{mid}$ and $m^{l}_{mid}$ represent the mid-frequency masks in the early and late denoising stages, respectively, where $m^{e}_{mid}$ covers a wider range of mid-frequency bands than $m^{l}_{mid}$.
To be analogous, $m^{e}_{low}$ and $m^{l}_{low}$ denote the low-frequency masks in the early and late denoising stages, respectively, with $m^{e}_{low}$ encompassing a broader range of low-frequency bands than $m^{l}_{low}$.
Notably, a larger ratio $\Delta_{cust}$ or $\Delta_{cost}$ indicates that the mixed reference of early stage contains more meaningful semantics than that within the late stage, hence more timesteps in the \textit{early} stage and fewer timesteps in the \textit{late} stage are required. For the timesteps of the denoising process from $T$ to 0, as per the threshold $\Delta \in \{\Delta_{cust}, \Delta_{cost}\}$ in Eq.~(\ref{eq:ratio}), we adaptively define the timesteps interval within $t \in [\frac{1}{1+\Delta}T, T]$ to cover the early stage of the denoising process, while the late stage as $t \in [0, \frac{1}{1+\Delta}T)$. \textit{\textbf{More details about the algorithm can be accessed from Sec.\ref{sec1} of the Appendix.}}


%% file: sec/4_experi.tex
\section{Experiments}
\subsection{Implementation Details} \label{sec:id}
We evaluate \textbf{Dual-FDM} on combined datasets covering image customization, style transfer, and text-to-image (T2I) generation: for small-scale customization-style transfer, we use \textbf{DreamBench} \cite{ruiz2023dreambooth} and \textbf{TI benchmark} \cite{gal2022image} as customization references with \textbf{IS-Plus} \cite{wang2024instantstyle1} as style references; for large-scale settings, we adopt \textbf{DreamBench++} \cite{pengdreambench++} and \textbf{StyleBench} \cite{gao2024styleshot}; and for color style transfer, we combine \textbf{IS-Plus} and \textbf{StyleBench} with prompts from \textbf{T2I-CompBench++} \cite{huang2025t2icompbench++}. We provide 9 evaluation metrics: (i) \textbf{Text and Image Alignment} via \textbf{CLIP}\footnote{Since CLIP is trained on real image–text (T) pairs, it is particularly sensitive to image style, which may distort text–image similarity. To alleviate this, we propose CLIP-TS by measuring similarity with style reference (S), thereby reducing the style influence . In addition, the generated background aligned with text prompts can interfere with the customized reference (C); hence we propose CLIP-TC, which incorporates text prompts to suppress such interference. We also provide CLIP-T (image–text similarity) and CLIP-CS (image–image reference similarity) for comparison.} \cite{radford2021learning} and \textbf{DINO} \cite{caron2021emerging}; (ii) \textbf{Image Quality} via \textbf{Image Reward (IR)} \cite{xu2023imagereward} and \textbf{HPS} \cite{wu2023human}; and (iii) \textbf{Efficiency} via inference time (\textbf{Times}) and GPU memory (\textbf{VRAM}). Our main experiments are conducted on the pre-trained IP-Adapter-Plus \cite{ye2023ip} with SDXL \cite{podell2023sdxl} model as the text-to-image diffusion model, configured with 30 sampling steps and and a fixed random seed of 42, to ensure high-quality outputs. We use the recommended hyper-parameters for each method across all images to ensure fair comparison. All experiments are implemented in PyTorch and conducted on a  NVIDIA RTX 4090 GPU. \textit{\textbf{More implementation details can be accessed in the Sec.\ref{sec3} of the Appendix.}}

\begin{table*}[t]
\caption{Quantitative comparisons between \textbf{Dual-FDM$_{cust}$} and other diffusion-based personalization models on several datasets.
\textcolor{red!20}{$\blacksquare$} and \textcolor{blue!20}{$\blacksquare$} denote the best and second-best results, respectively.
\textbf{Dual-FDM$_{cust}$} consistently achieves the best performance.}
\label{tab:performance_all}
\centering
\renewcommand{\arraystretch}{0.9}
\begin{adjustbox}{max width=\textwidth}
\begin{tabular}{@{}cl *{7}{c}@{}}
\toprule[1.5pt]
\multicolumn{2}{c}{\textbf{Baseline}}
& \makecell[c]{\textbf{IP-Adapter} \\ Arxiv'23} &
  \makecell[c]{\textbf{SSR-Encoder} \\ CVPR'24} &
  \makecell[c]{\textbf{MIP-Adapter} \\ AAAI'25} &
  \makecell[c]{\textbf{Dreamcache} \\ CVPR'25} &
  \makecell[c]{\textbf{PatchDPO} \\ CVPR'25} &
  \makecell[c]{\textbf{MS-Diffusion} \\ ICLR'25} &
  \textsc{\textbf{Dual-FDM$_{cust}$}}\\
\midrule
\multirow{6}{*}[-0.2em]{%
  \raisebox{0.0em}[0pt][0pt]{%
    \rotatebox[origin=c]{90}{%
      \parbox[c]{3.2cm}{\centering\textsc{\textbf{CSGO} \\ (Arxiv'24)}}
    }%
  }%
}
& CLIP-T $\uparrow$ & 0.2851 & 0.2829 & 0.2809 & 0.2803 & 0.2793 & \cellcolor{blue!20}0.3123 & \cellcolor{red!20}0.3194 \\
& CLIP-TC $\uparrow$ & \cellcolor{blue!20}0.4802 & 0.4689 & 0.4744 & 0.4518 & 0.4738 & 0.4680 & \cellcolor{red!20}0.5031 \\
& CLIP-TS $\uparrow$ & 0.4415 & 0.4439 & 0.4420 & 0.4525 & 0.4440 & \cellcolor{red!20}0.4577 & \cellcolor{blue!20}0.4516 \\
& CLIP-CS $\uparrow$ & \cellcolor{blue!20}0.6658 & 0.6587 & 0.6626 & 0.6465 & 0.6638 & 0.6419 & \cellcolor{red!20}0.6687 \\
& IR$_{\times 10}$ $\uparrow$ & -5.5733 & -7.4241 & -6.5255 & -6.7782 & -7.1456 & \cellcolor{blue!20}4.2023 & \cellcolor{red!20}5.3487 \\
& HPS$_{\times 10}$ $\uparrow$ & 2.3401 & 2.2692 & 2.2624 & 2.1794 & 2.2493 & \cellcolor{blue!20}2.5436 & \cellcolor{red!20}2.6007 \\
\midrule

\multirow{6}{*}[-0.2em]{%
  \raisebox{0.0em}[0pt][0pt]{%
    \rotatebox[origin=c]{90}{%
      \parbox[c]{3.2cm}{\centering\textsc{\textbf{StyleID}\\ (CVPR'24)}}
    }%
  }%
}
& CLIP-T $\uparrow$ & 0.2731 & 0.2768 & \cellcolor{blue!20}0.2848 & 0.2647 & 0.2650 & \cellcolor{blue!20}0.3012 & \cellcolor{red!20}0.3194 \\
& CLIP-TC $\uparrow$ & \cellcolor{blue!20}0.4828 & 0.4773 & 0.4699 & 0.4424 & 0.4744 & 0.4614 & \cellcolor{red!20}0.5031 \\
& CLIP-TS $\uparrow$ & 0.4349 & 0.4370 & \cellcolor{blue!20}0.4458 & 0.4488 & 0.4370 & \cellcolor{red!20}0.4556 & \cellcolor{blue!20}0.4516 \\
& CLIP-CS $\uparrow$ & \cellcolor{blue!20}0.6731 & 0.6665 & 0.6584 & 0.6460 & 0.6710 & 0.6423 & \cellcolor{red!20}0.6687 \\
& IR$_{\times 10}$ $\uparrow$ & -8.6015 & -8.7602 & -9.5626 & -9.9445 & -10.4733 & \cellcolor{blue!20}2.6485 & \cellcolor{red!20}5.3487 \\
& HPS$_{\times 10}$ $\uparrow$ & 2.0861 & 2.0979 & 2.0073 & 1.9126 & 1.9978 & \cellcolor{blue!20}2.3322 & \cellcolor{red!20}2.6007 \\
\midrule


\multirow{6}{*}[-0.2em]{%
  \raisebox{0.0em}[0pt][0pt]{%
    \rotatebox[origin=c]{90}{%
      \parbox[c]{3.2cm}{\centering\textsc{\textbf{StyleSSP}\\ (CVPR'25)}}
    }%
  }%
}
& CLIP-T $\uparrow$ & 0.2642 & 0.2656 & 0.2584 & 0.2561 & 0.2591 & \cellcolor{blue!20}0.2907 & \cellcolor{red!20}0.3194 \\
& CLIP-TC $\uparrow$ & 0.4664 & \cellcolor{blue!20}0.4729 & 0.4616 & 0.4355 & 0.4638 & 0.4540 & \cellcolor{red!20}0.5031 \\
& CLIP-TS $\uparrow$ & 0.4397 & 0.4407 & 0.4409 & 0.4468 & 0.4413 & \cellcolor{red!20}0.4583 & \cellcolor{blue!20}0.4516 \\
& CLIP-CS $\uparrow$ & 0.6680 & 0.6592 & 0.6678 & 0.6433 & \cellcolor{blue!20}0.6700 & 0.6462 & \cellcolor{red!20}0.6687 \\
& IR$_{\times 10}$ $\uparrow$ & -10.8440 & -10.4537 & -12.0747 & -11.0862 & -12.1339 & \cellcolor{blue!20}0.5184 & \cellcolor{red!20}5.3487 \\
& HPS$_{\times 10}$ $\uparrow$ & 2.0087 & 2.0153 & 1.9075 & 1.8562 & 1.9204 & \cellcolor{blue!20}2.2367 & \cellcolor{red!20}2.6007 \\
\bottomrule[1.5pt]
\end{tabular}
\end{adjustbox}
\end{table*}

\begin{figure*}[t]
  \centering
  \includegraphics[width=\textwidth]{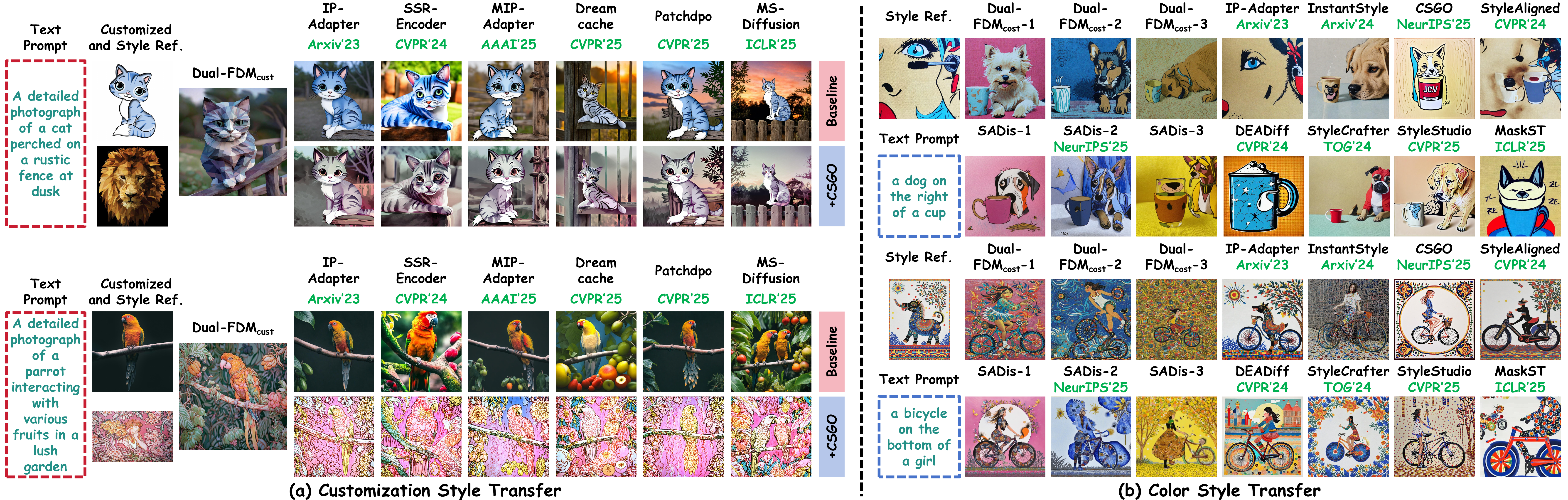}
  \caption{Comparison of (a) customization style transfer and (b) color style  transfer with state-of-the-arts. \textbf{Dual-FDM} delivers superior results compared with existing approaches.}
  \label{fig:compare}
\end{figure*}

\subsection{Comparison with State-of-the-arts}\label{sec:com}
\subsubsection{Quantitative and Qualitative Analysis for Customization Style Transfer}
To validate the superiority of \textbf{Dual-FDM$_{cust}$}, we perform a thorough comparison against state-of-the-art diffusion models for customization-style transfer. For fair comparison, we also compare with both image customization and style transfer. For image customization, we consider two groups of representative methods. The first group includes \textbf{IP-Adapter} \cite{ye2023ip}, \textbf{Dreamcache} \cite{aiello2025dreamcache}, and \textbf{PatchDPO} \cite{huang2025resolving}, while these approaches  emphasize dense image references of the foreground while overwhelming sparse text prompts of the background, especially for complex ones. The second group consists of \textbf{SSR-Encoder} \cite{zhang2024ssr}, \textbf{MIP-Adapter} \cite{huang2025resolving}, and \textbf{MS-Diffusion} \cite{wangms}, resorting to layout constraints to integrate customized references. This preserves the foreground identity, but the explicit separation of foreground and background in the layout disrupts overall consistency, leading to misalignment with (complex / long) text prompts. Building on the above, we adopt the latest image-driven style transfer methods for the second stage of the style transfer, including \textbf{CSGO} \cite{xing2024csgo}, \textbf{StyleID} \cite{Chung_2024_CVPR},  \textbf{StyleSSP} \cite{xu2025stylessp}.

The quantitative results in Table.~\ref{tab:performance_all} summarize our findings below: \textbf{Dual-FDM$_{cust}$} enjoys larger CLIP-T, CLIP-TC, CLIP-TS, CLIP-CS, IR and HPS than the competitors. Notably, [\textbf{MS-Diffusion} \cite{wangms} + \textbf{CSGO} \cite{xing2024csgo}] remains the large performance margins at most 2.27\% for CLIP-T, \ 7.5\% for CLIP-TC,  \ 4.17\% for CLIP-CS  \ 27.28\% for DINO-CS and \ 2.24\% for HPS) compared to \textbf{Dual-FDM$_{cust}$} in Table.~\ref{tab:performance_all}, the results verifies the intuition in Sec.~\ref{intord} -- \textit{the dominance of low-and-mid over high frequency bands—especially when entangled within a single image reference—tends to reinforce each other at the early stages, resulting in an overwhelming suppression of text prompt guidance.} We also illustrate more generation results in Fig.~\ref{fig:compare}(a).  \textbf{\textit{More analysis of computational efficiency can be seen in Sec.\ref{sec4_0} of the Appendix.}}


\begin{table}
\caption{Quantitative comparisons for color style transfer between \textbf{Dual-FDM$_{cost}$} and other diffusion-based personalization models on T2I-CompBench++. \textcolor{red!20}{$\blacksquare$} and \textcolor{blue!20}{$\blacksquare$} denote the best and second-best results.
\textbf{Dual-FDM$_{cost}$} achieves the best performance. }
\centering
\label{tab:performance_all1}
\small
\setlength{\tabcolsep}{4pt}
\begin{adjustbox}{max width=0.96\linewidth}
\begin{tabular}{@{}cc*{9}{c}@{}}
\toprule[1.5pt]
 & &
\rotatebox{0}{\textbf{InstantStyle}} &  
\rotatebox{0}{\textbf{CSGO}} &        
\rotatebox{0}{\textbf{StyleShot}} &   
\rotatebox{0}{\textbf{StyleAligned}} & 
\rotatebox{0}{\textbf{DEADiff}} &
\rotatebox{0}{\textbf{StyleCrafter}} &
\rotatebox{0}{\textbf{MaskST}} &
\rotatebox{0}{\textbf{StyleStudio}} &
\rotatebox{0}{\multirow{2}{*}{\textbf{Dual-FDM$_{cost}$}}} \\
& & \scriptsize Arxiv'24 & \scriptsize NeurIPS'25 & \scriptsize Arxiv'24 & \scriptsize CVPR'24 & \scriptsize CVPR'24 & \scriptsize TOG'24 & \scriptsize ICLR'25 & \scriptsize CVPR'25 & \\
\midrule[1pt]

\multirow{3}{*}[0.2em]{%
  \raisebox{0.0em}[0pt][0pt]{%
    \rotatebox[origin=c]{90}{%
      \parbox[c]{3.2cm}{\centering Color}
    }%
  }%
}
& CLIP-TS $\uparrow$ & 0.4587 & 0.4539 & 0.4644 & \cellcolor{blue!20}0.4832 & 0.4448 & 0.4771 & 0.4641 & 0.4497 & \cellcolor{red!20}0.5015 \\
& CLIP-S $\uparrow$ & 0.5967 & 0.6392 & 0.6782 & \cellcolor{blue!20}0.6904 & 0.6177 & 0.6304 & 0.6763 & 0.5859 & \cellcolor{red!20}0.7178 \\
& DINO-S $\uparrow$ & 0.2146 & 0.3924 & 0.4769 & 0.4437 & 0.3318 & 0.3541 & \cellcolor{blue!20}0.4764 & 0.2370 & \cellcolor{red!20}0.4970 \\
\cmidrule[0.5pt](lr){2-11}

\multirow{3}{*}[0.2em]{%
  \raisebox{0.0em}[0pt][0pt]{%
    \rotatebox[origin=c]{90}{%
      \parbox[c]{3.2cm}{\centering Shape}
    }%
  }%
}
& CLIP-TS $\uparrow$ & 0.4507 & 0.4561 & 0.4595 & \cellcolor{blue!20}0.4797 & 0.4392 & 0.4656 & 0.4597 & 0.4504 & \cellcolor{red!20}0.4956 \\
& CLIP-S $\uparrow$ & 0.5864 & 0.6357 & 0.6602 & \cellcolor{blue!20}0.7051 & 0.5986 & 0.6274 & 0.6602 & 0.6162 & \cellcolor{red!20}0.7271 \\
& DINO-S $\uparrow$ & 0.2087 & 0.3711 & 0.4374 & \cellcolor{blue!20}0.4658 & 0.2944 & 0.3843 & 0.4368 & 0.3087 & \cellcolor{red!20}0.4982 \\
\cmidrule[0.5pt](lr){2-11}

\multirow{3}{*}[0.2em]{%
  \raisebox{0.0em}[0pt][0pt]{%
    \rotatebox[origin=c]{90}{%
      \parbox[c]{3.2cm}{\centering Texture}
    }%
  }%
}
& CLIP-TS $\uparrow$ & 0.4524 & 0.4561 & 0.4543 & \cellcolor{blue!20}0.4819 & 0.4304 & 0.4712 & 0.4534 & 0.4443 & \cellcolor{red!20}0.4988 \\
& CLIP-S $\uparrow$ & 0.5879 & 0.6406 & 0.6553 & \cellcolor{blue!20}0.7246 & 0.5835 & 0.6348 & 0.6538 & 0.5889 & \cellcolor{red!20}0.7358 \\
& DINO-S $\uparrow$ & 0.2131 & 0.3760 & 0.4476 & \cellcolor{blue!20}0.4844 & 0.2707 & 0.3729 & 0.4465 & 0.2616 & \cellcolor{red!20}0.4991 \\
\cmidrule[0.5pt](lr){1-11}

\multirow{3}{*}[-0.2em]{%
  \raisebox{0.0em}[0pt][0pt]{%
    \rotatebox[origin=c]{90}{%
      \parbox[c]{3.2cm}{\centering Non \\ Spatial}
    }%
  }%
}
& CLIP-TS $\uparrow$ & 0.4424 & 0.4553 & 0.4504 &  \cellcolor{blue!20}0.4839 & 0.4402 & 0.4553 & 0.4504 & 0.4314 & \cellcolor{red!20}0.4958 \\
& CLIP-S $\uparrow$ & 0.5723 & 0.6279 & 0.6387 &  \cellcolor{blue!20}0.7251 & 0.5972 & 0.6064 & 0.6387 & 0.5566 & \cellcolor{red!20}0.7261\\
& DINO-S $\uparrow$ & 0.2336 & 0.4278 & 0.4853 &  \cellcolor{blue!20}0.5171 & 0.3533 & 0.3795 & 0.4854 & 0.2646 & \cellcolor{red!20}0.5406 \\
\cmidrule[0.5pt](lr){2-11}

\multirow{3}{*}[-0.2em]{%
  \raisebox{0.0em}[0pt][0pt]{%
    \rotatebox[origin=c]{90}{%
      \parbox[c]{3.2cm}{\centering 2D \\ Spatial}
    }%
  }%
}
& CLIP-TS $\uparrow$ & 0.4568 & 0.4670 & 0.4661 & \cellcolor{blue!20}0.4912 & 0.4563 & 0.4829 & 0.4668 & 0.4565 & \cellcolor{red!20}0.5122 \\
& CLIP-S $\uparrow$ & 0.5835 & 0.6309 & 0.6592 & \cellcolor{blue!20}0.7129 & 0.6362 & 0.6421 & 0.6597 & 0.6060 & \cellcolor{red!20}0.7314 \\
& DINO-S $\uparrow$ & 0.2262 & 0.4113 & 0.4615 & \cellcolor{blue!20}0.4998 & 0.3824 & 0.3977 & 0.4647 & 0.3253 & \cellcolor{red!20}0.5187 \\
\cmidrule[0.5pt](lr){2-11}

\multirow{3}{*}[-0.2em]{%
  \raisebox{0.0em}[0pt][0pt]{%
    \rotatebox[origin=c]{90}{%
      \parbox[c]{3.2cm}{\centering 3D \\ Spatial}
    }%
  }%
}
& CLIP-TS $\uparrow$ & 0.4573 & 0.4624 & 0.4663 & \cellcolor{blue!20}0.4875 & 0.4497 & 0.4773 & 0.4663 & 0.4429 & \cellcolor{red!20}0.5078 \\
& CLIP-S $\uparrow$ & 0.5918 & 0.6333 & 0.6680 & \cellcolor{blue!20}0.7100 & 0.6333 & 0.6357 & 0.6680 & 0.5732 & \cellcolor{red!20}0.7334 \\
& DINO-S $\uparrow$ & 0.2166 & 0.3819 & 0.4654 & \cellcolor{blue!20}0.4760 & 0.3655 & 0.3774 & 0.4696 & 0.2249 & \cellcolor{red!20}0.5094 \\
\cmidrule[0.5pt](lr){1-11}
\multirow{3}{*}[0.3em]{%
  \raisebox{0.0em}[0pt][0pt]{%
    \rotatebox[origin=c]{90}{%
      \parbox[c]{3.2cm}{\centering Number}
    }%
  }%
}
& CLIP-TS $\uparrow$ & 0.4490 & 0.4509 & 0.4597 & \cellcolor{blue!20}0.4902 & 0.4465 & 0.4783 & 0.4597 & 0.4436 & \cellcolor{red!20}0.5029 \\
& CLIP-S $\uparrow$ & 0.5845 & 0.6182 & 0.6562 & \cellcolor{blue!20}0.7241 & 0.6226 & 0.6499 & 0.6558 & 0.6011 & \cellcolor{red!20}0.7334 \\
& DINO-S $\uparrow$ & 0.2289 & 0.3882 & 0.4779 & \cellcolor{blue!20}0.5116 & 0.3541 & 0.4346 & 0.4738 & 0.3160 & \cellcolor{red!20}0.5469 \\
\cmidrule[0.5pt](lr){1-11}
\multirow{3}{*}[0.3em]{%
  \raisebox{0.0em}[0pt][0pt]{%
    \rotatebox[origin=c]{90}{%
      \parbox[c]{3.2cm}{\centering Complex}
    }%
  }%
}
& CLIP-TS $\uparrow$ & 0.4404 & 0.4500 & 0.4556 & \cellcolor{blue!20}0.4836 & 0.4390 & 0.4636 & 0.4558 & 0.4351 & \cellcolor{red!20}0.4954 \\
& CLIP-S $\uparrow$ & 0.5864 & 0.6445 & 0.6582 & \cellcolor{blue!20}0.7256 & 0.6021 & 0.6387 & 0.6582 & 0.5791 & \cellcolor{red!20}0.7329 \\
& DINO-S $\uparrow$ & 0.2134 & 0.3911 & 0.4533 & \cellcolor{blue!20}0.4961 & 0.2938 & 0.3956 & 0.4536 & 0.2274 & \cellcolor{red!20}0.5051 \\

\bottomrule[1.5pt]
\end{tabular}
\end{adjustbox}
\end{table}

\subsubsection{Quantitative and Qualitative Analysis for Color Style Transfer}
we conduct additional evaluations against state-of-the-art diffusion models for color style transfer. These include \textbf{InstantStyle} \cite{wang2024instantstyle}, \textbf{CSGO} \cite{xing2024csgo}, \textbf{StyleShot} \cite{gao2024styleshot}, \textbf{StyleAligned} \cite{hertz2024style}, \textbf{DEADiff} \cite{qi2024deadiff}, \textbf{StyleCrafter} \cite{liu2023stylecrafter}, \textbf{MaskST} \cite{zhuless}, \textbf{IP-Adapter} \cite{ye2023ip}, and \textbf{SADis} \cite{qin2025freelunchcolortexturedisentanglementstylized}. However, these methods often suffer from style leakage, leading to misalignment with text prompts. \textbf{StyleStudio} \cite{lei2025stylestudio} employs layout to preserve the foreground, but it still struggles to handle text prompts involving multiple objects, \textbf{SADis} \cite{qin2025freelunchcolortexturedisentanglementstylized} mainly focuses on color style transfer, but it often fails to capture enough stylistic details from the reference due to entanglement across frequency bands.

We employ more complex text prompts in \textbf{T2I-CompBench++} to generate the style transfer result in Table.\ref{tab:performance_all1}, which indicates that \textbf{Dual-FDM$_{cost}$} enjoys larger CLIP-TS, CLIP-S and DINO-S  than the competitors in different kinds of  compositional text prompts. Notably, Our \textbf{Dual-FDM$_{cost}$} maintains achieves the large performance gain over \textbf{StyleAligned}, which indicates that \textbf{Dual-FDM$_{cost}$} outperforms by \textit{substitute the low-frequency band of the background in the style reference with that from both the foreground and background of the color reference} (See Sec.~\ref{sec:ctst}). To shed more light on the advantages of our method, we further perform the generation analysis on the color style transfer results as illustrated in Fig.~\ref{fig:compare}(b). \textbf{\textit{More generation results for different T2I models can be seen in Sec.\ref{sec4} of the Appendix.}}

\subsection{Ablation Studies} \label{sec:ab}
\subsubsection{Discussion on Different Modules of Dual-FDM.}
To validate of our \textbf{Dual-FDM$_{cust}$}, we perform the ablation study on DreamBooth/TI benchmark and DreamBooth++ with two variants:  \textbf{Case A}: removing  $\overline{W}_k^{im}$ in Eq.~(\ref{eq:style});  \textbf{Case B}: setting  the same mid-frequency masks for the early and late denoising stages. Table.~\ref{tab:ab1} reports that our \textbf{Dual-FDM$_{cust}$} exhibits the higher similarity to the {style reference} (CLIP-TS and CLIP-S)  than \textbf{Case A}, confirming that \textit{$\overline{W}_k^{im}$ can enhance the background of the style reference}, which in line with Sec.~\ref{sec:back}. \textbf{Case B} exhibits a suboptimal performance degradation, confirming that \textit{the mid-frequency band converge more rapidly during the early stages of the denoising process under high-level noise, while the high-frequency band evolves more gradually}, confirming the intuition of Sec.~\ref{intord}, along with the intuition of Fig.~\ref{fig:ab2}.

\subsubsection{Discussion on Denoising Timesteps Between the Early and Late Stages.}
\begin{figure*}
\centering

\begin{minipage}{0.48\textwidth}
    \centering
    
    \includegraphics[width=0.9\linewidth]{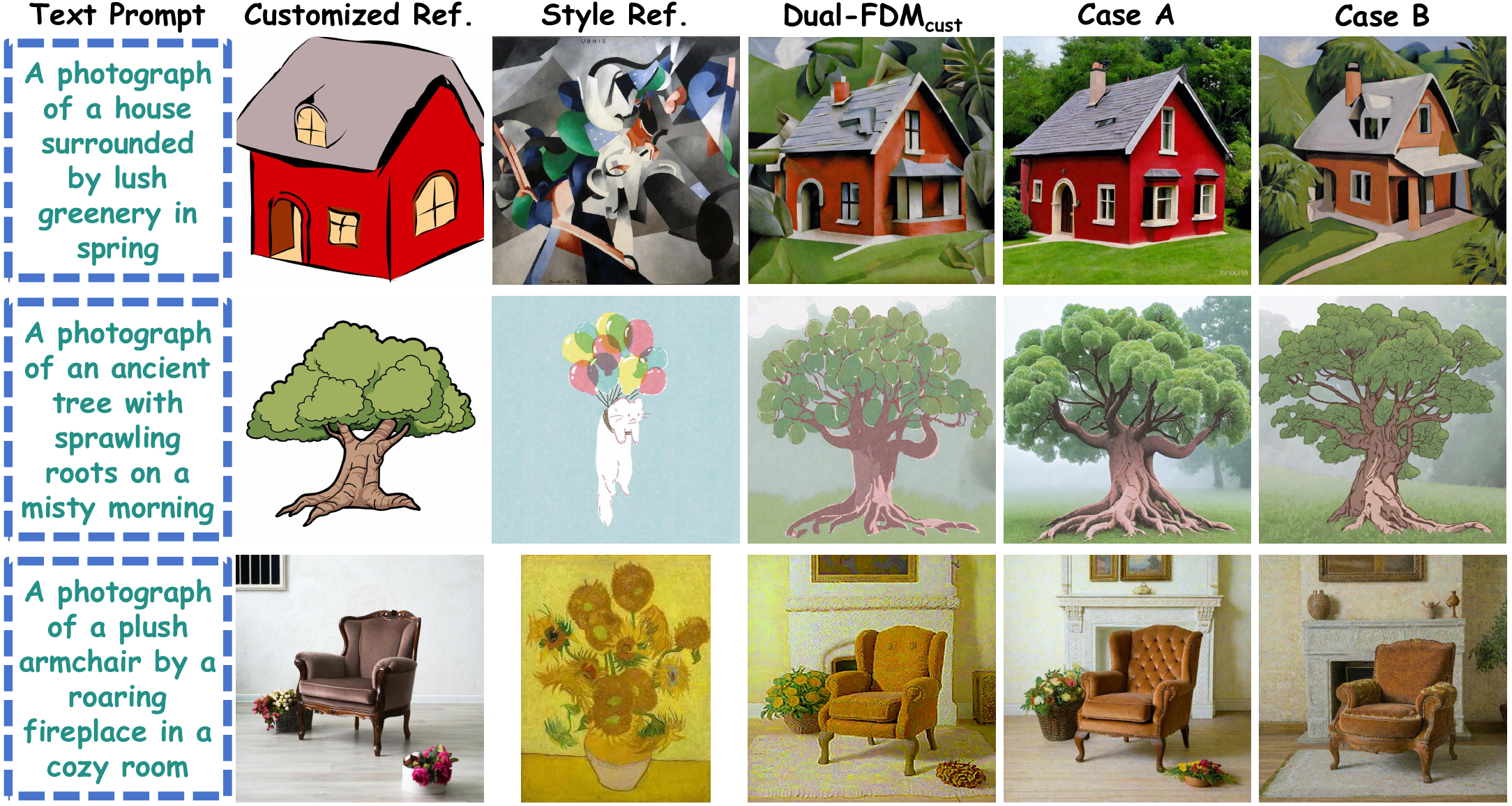}
    \captionof{figure}{Ablation studies about the influence of different modules on generating images.}
    \label{fig:ab2}
    
    \vspace{0.5em}
    
    \captionof{table}{Ablation studies on the influence of different modules on image generation. 
    \textbf{Case A}: removing $\overline{W}_k^{im}$ in Eq.~(\ref{eq:style}); 
    \textbf{Case B}: using the same mid-frequency masks for early and late denoising stages. 
    \textcolor{red!20}{$\blacksquare$} denotes the best results.}
    \resizebox{\linewidth}{!}{
    \begin{tabular}{c|cccc||cccc}
    \toprule[1.5pt]
    \multirow{2}{*}{Method} 
    & \multicolumn{4}{c||}{\textbf{DreamBooth/TI benchmark}} 
    & \multicolumn{4}{c}{\textbf{DreamBooth++}} \\
    
    & CLIP-TC$\uparrow$ & CLIP-TS$\uparrow$ & CLIP-C$\uparrow$ & CLIP-S$\uparrow$  
    & CLIP-TC$\uparrow$ & CLIP-TS$\uparrow$ & CLIP-C$\uparrow$ & CLIP-S$\uparrow$ \\
    
    \midrule
    Case A 
    & \cellcolor{red!20}0.5171 & 0.4346 & \cellcolor{red!20}0.7817 & 0.5215
    & \cellcolor{red!20}0.5562 & 0.4395 & \cellcolor{red!20}0.8438 & 0.5459 \\
    
    Case B 
    & 0.5039 & 0.4387 & 0.7510 & 0.5254
    & 0.5444 & 0.4512 & 0.8042 & 0.5547 \\
    
    \textbf{Dual-FDM$_{cust}$} 
    & 0.5054 & \cellcolor{red!20}0.4468 & 0.7676 & \cellcolor{red!20}0.5571
    & 0.5459 & \cellcolor{red!20}0.4526 & 0.8208 & \cellcolor{red!20}0.5737 \\
    
    \bottomrule[1.5pt]
    \end{tabular}}
    
    \label{tab:ab1}

\end{minipage}
\hfill
\begin{minipage}{0.48\textwidth}
    \centering
    \includegraphics[width=\linewidth]{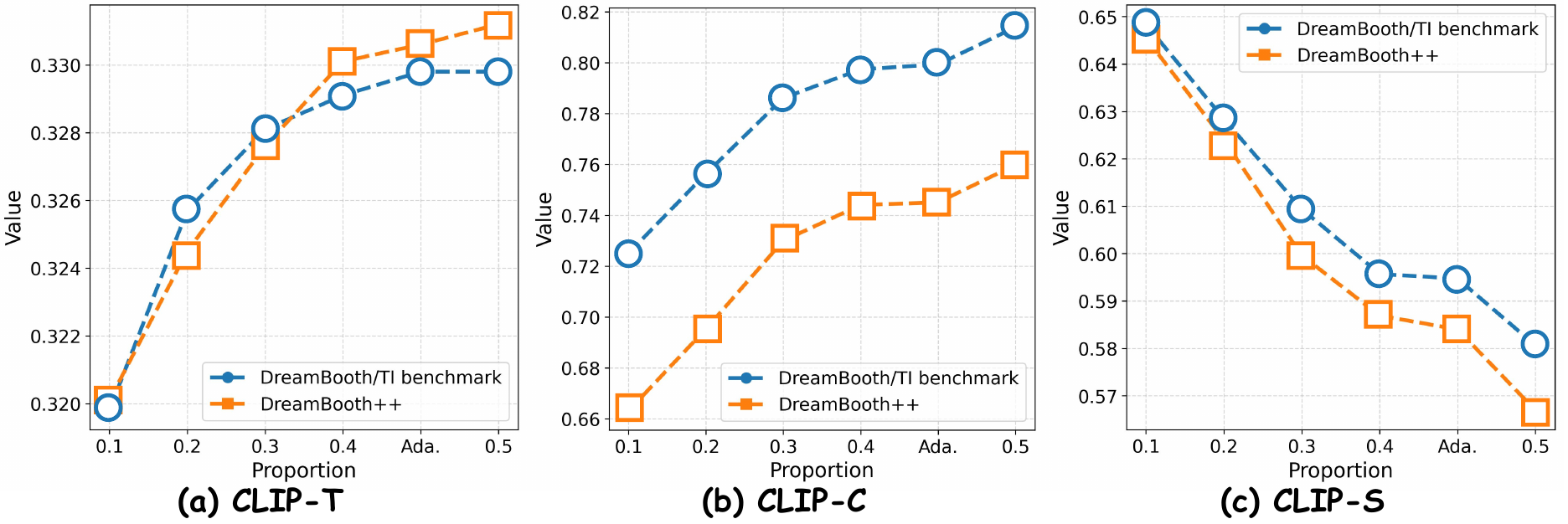}
    \captionof{figure}{Ablation studies about the influence of different denoising timesteps between the early and late stages on generating the images.}
    \label{fig:ab1}
    
    \vspace{1em}
    
    \captionof{table}{Ablation studies on the effect of masks filtering mid- and low-frequency bands for image generation. 
    \textcolor{red!20}{$\blacksquare$} denotes the best results.}
    \resizebox{\linewidth}{!}{
    \begin{tabular}{c|c|cccccc}
    \toprule[1.5pt]
    Frequency Band & Case & CLIP-T$\uparrow$ & CLIP-C$\uparrow$ & DINO-C$\uparrow$ & CLIP-S$\uparrow$ & DINO-S$\uparrow$ \\
    
    \cmidrule(lr){1-7}
    \multirow{4}{*}{Mid-Frequency} 
    & $th_{mp1}=0.1, th_{mp2}=0.9$ & 0.3313 & \cellcolor{red!20}0.8628 & \cellcolor{red!20}0.7082 & 0.5249 & 0.1239 \\
    
    & $th_{mp1}=0.2, th_{mp2}=0.8$ & 0.3425 & 0.8247 & 0.6168 & 0.5420 & 0.1564 \\
    
    & $th_{mp1}=0.3, th_{mp2}=0.7$ & \cellcolor{red!20}0.3467 & 0.7798 & 0.5050 & 0.5723 & 0.2177 \\
    
    & $th_{mp1}=0.4, th_{mp2}=0.6$ & 0.3311 & 0.7236 & 0.3909 & \cellcolor{red!20}0.6348 & \cellcolor{red!20}0.3206 \\
    
    \midrule
    \midrule
    
    Frequency Band  & Case & CLIP-T$\uparrow$ & CLIP-S$\uparrow$ & DINO-S$\uparrow$ & $Color_{l_{2}}$$\downarrow$ &$Color_{l_{1}}$$\downarrow$ \\
    
    \cmidrule(lr){1-7}
    \multirow{4}{*}{Low-Frequency} 
    & $th_{lp}=0.8$ & 0.4304 & 0.5420 & 0.1069 & \cellcolor{red!20}5.4735 & \cellcolor{red!20}1.3168 \\
    
    & $th_{lp}=0.6$ & 0.4419 & 0.5645 & 0.1550 & 5.6028 & 1.3453 \\
    
    & $th_{lp}=0.4$ & 0.4519 & 0.5928 & 0.2200 & 6.0283 & 1.3775 \\
    
    & $th_{lp}=0.2$ & \cellcolor{red!20}0.4692 & \cellcolor{red!20}0.6372 & \cellcolor{red!20}0.3070 & 6.2901 & 1.3969 \\
    
    \bottomrule[1.5pt]
    \end{tabular}}
    
    \label{tab:ab3}

\end{minipage}
\end{figure*}

We conduct ablation studies by dividing the denoising process into early and late stages according to different masks of mid-frequency bands, \textit{i.e.}, $m^{e}_{mid}$ and $m^{l}_{mid}$, and perform experiments by varying the proportion of the early stage, \textit{i.e.}, \{0.1; 0.2; 0.3; 0.4; 0.5\}, We also provide the result of adaptively modulating the early and late denoising stage, name \textbf{Ada.}.  Fig.~\ref{fig:ab1} reports the results for all cases, When the proportion is set to 0.1, the generated images show the highest similarity to the style reference while being least similar to the customized reference; conversely, at 0.5, the opposite trend is observed. We can also conclude that \textbf{Ada.} lies between the results of the 0.4 and 0.5 proportions, demonstrating that our dynamic adjustment strategy can effectively regulate the ratio of early to late stages, leading to a well-balanced performance between text alignment and image similarity, confirming the intuition of Sec.~\ref{sec:ada}. \textbf{\textit{More generation results for dual references can be seen in Sec.\ref{sec5} of the Appendix.}}

\subsubsection{Discussion on Different Masks to filter Mid-and-Low Frequency Bands.}
We conduct ablation studies on \textbf{Dual-FDM$_{cust}$} and \textbf{Dual-FDM$_{cost}$} to investigate the effects of different frequency bands. Specifically, for the mid-frequency band, we define different thresholds $(th_{mp1}, th_{mp2})$, where $(th_{mp1}=0.1, th_{mp2}=0.9)$ represents the widest mid-frequency range and $([0.4, 0.6]$ represents a narrower range. For the low-frequency band, we vary $th_{lp}\in\{0.9,0.8,0.7,0.6\}$, where $th_{lp}=0.9$ indicates the largest proportion of low-frequency components. Table.~\ref{tab:ab3} reports the results under different configurations. When the mid-frequency range is the widest (\textit{i.e.}, $[0.1, 0.9]$), the generated images show the highest similarity to the {customized reference} (as reflected by CLIP-C and DINO-C) and the lowest similarity to the {style reference} (CLIP-S and DINO-S). As the mid-frequency range becomes narrower (\textit{i.e.}, $th_{mp1}=0.4, th_{mp2}=0.6$), the opposite trend is observed, suggesting that a narrower mid-frequency band enhances the representation of style features. For the low-frequency band, where lower $Color_{l_{2}}$ and $Color_{l_{1}}$ values indicate better color consistency, decreasing $th_{lp}$ from 0.8 to 0.2 (thus reducing the proportion of low-frequency bands) leads to higher similarity between the generated images and the style reference (CLIP-S, CLIP-TS, DINO-S), but worsens color consistency — the color error metrics ($Color_{l_{2}}$, $Color_{l_{1}}$) increase, which is consistent with Sec.~\ref{sec:ctst}. \textbf{\textit{More generation results for different masks can be accessed from Sec.\ref{sec6} of the Appendix.}}

%% file: sec/5_appendix.tex
Due to page limitation of the mainbody, the supplementary material offers further more quantitative results and qualitative results with higher resolution, which are summarized below:
\begin{itemize}[leftmargin=2em]
    \item {The frequency variation during the denoising process, as mentioned in Sec.~\ref{intord} of the mainbody (Sec.~\ref{sec1s})}
    \item More intuitions for the rationale between ${W}_k^{im}$ and $\overline{W}_k^{im}$, as mentioned in Sec.2.2 of the mainbody (Sec.~\ref{sec2})
    \item The algorithm of customization style transfer and  color style transfer, as mentioned in Sec.2 of the mainbody (Sec.~\ref{sec1})
    \item More details about the implementation details, as mentioned in Sec.3.1 of the mainbody. (Sec.~\ref{sec3})
    \item More Analysis of Computational Efficiency in the Sec.3.2 of the Mainbody. (Sec.~\ref{sec4_0})
    \item More generation results of different T2I models, as mentioned in Sec.3.2 of the mainbody. (Sec.~\ref{sec4})
    \item Additional generation results for the ablation study about the denoising timesteps between the early and late stages, as mentioned in Sec.3.3 of the mainbody. (Sec.~\ref{sec5})
    \item Additional generation results for the ablation study about the different masks to filter mid-and-low frequency bands, as mentioned in Sec.3.3 of the mainbody. (Sec.~\ref{sec6})
    \item We list the limitations and broader impacts (Sec.~\ref{secli}).
    \item Discussion on Additional Metrics and Style-Transfer Baseline (Sec.~\ref{sec:style_transfer_baselines}).
    \item More Results on Extreme Scenarios (Sec.~\ref{sec:extreme_scenarios}).
    \item Related work (Sec.~\ref{rel}).
\end{itemize}

\section{Frequency Variation During the Denoising Process}\label{sec1s}

{As mentioned in Sec.~\ref{intord} of the mainbody, We  claim that \textbf{during the denoising process, images recover low-and-mid frequency bands first, followed by high-frequency details}. {To validate the claims via frequency analysis, we experimentally measure the changes in both low- and high-frequency bands at different timesteps during the denoising process. The results are reported in Tab.~\ref{tab:freq_analysis}:}
\begin{itemize}[leftmargin=*, topsep=0pt, parsep=0pt]
    \item {\textbf{Low-frequency information.} We adopt \textit{Mean Gray}, \textit{Mean HSV-V}, and \textit{Mean Luma}, which primarily capture global brightness and luminance structure. These metrics quantify the stability of the low-frequency bands throughout the denoising process.}
    \item {\textbf{High-frequency information.} We use \textit{Average Gradient}, \textit{Variance}, and \textit{LBP Variance}, which are sensitive to edge sharpness, local contrast, and texture complexity. These metrics reflect the recovery of high-frequency details.}
\end{itemize}
{The low-frequency metrics remain nearly unchanged from the beginning (e.g., \textbf{Mean Gray varies only slightly from 119.5 to 120.9}). In contrast, the high-frequency metrics increase steadily throughout the reverse process (e.g., \textbf{Avg Gradient rises from 59.01 to 233.88}), confirming that high-frequency details are refined gradually during denoising. This is consistent with \textbf{Sec.~\ref{intord} of the mainbody}.}

\begin{table}[h]
\centering
\caption{The high-and-low frequency bands across different timesteps during the denoising process. \textcolor{red!20}{$\blacksquare$} denotes the best result.}
\label{tab:freq_analysis}
\begin{adjustbox}{max width=\textwidth}
\begin{tabular}{lcccccc}
\toprule
\textbf{Timesteps} & \textbf{Mean Gray} & \textbf{Mean HSV-V} & \textbf{Mean Luma} & \textbf{Avg Gradient} & \textbf{Variance} & \textbf{LBP Variance} \\
\midrule
\textbf{0}  & \cellcolor{red!20}{120.9} & \cellcolor{red!20}{136.0} & \cellcolor{red!20}{120.9} & \cellcolor{red!20}{59.01}  & \cellcolor{red!20}{3955.0} & \cellcolor{red!20}{4.45} \\
5   & 120.8 & 134.6 & 120.8 & 75.07  & 4098.3 & 3.71 \\
10  & 120.7 & 133.5 & 120.7 & 101.58 & 4378.9 & 3.25 \\
15  & 120.7 & 133.7 & 120.7 & 133.26 & 4790.8 & 3.07 \\
20  & 120.9 & 135.1 & 120.9 & 164.23 & 5274.8 & 3.07 \\
25  & 121.2 & 137.3 & 121.2 & 190.49 & 5743.9 & 3.15 \\
30  & 121.4 & 139.2 & 121.4 & 209.84 & 6099.5 & 3.25 \\
35  & 121.1 & 140.0 & 121.1 & 222.03 & 6291.1 & 3.32 \\
40  & 120.5 & 140.1 & 120.5 & 228.75 & 6363.9 & 3.35 \\
45  & 119.9 & 139.8 & 119.9 & 232.19 & 6394.8 & 3.38 \\
50  & 119.5 & 139.5 & 119.5 & 233.88 & 6419.5 & 3.40 \\
\bottomrule
\end{tabular}
\end{adjustbox}
\end{table}
\section{More intuitions for the rationale between ${W}_k^{im}$ and $\overline{W}_k^{im}$}\label{sec2}
As indicated in Sec.2.2 of the mainbody, we provide more intuitions for the rationale between ${W}_k^{im}$ and $\overline{W}_k^{im}$. To verify the intuition that \textbf{larger $\sigma_i$ tends to correspond to the foreground, whereas smaller $\sigma_i$ is more likely associated with the background}, as shown in Tab.~\ref{table:intuitions},
we utilize the text prompts for both the foreground and background to generate their corresponding images using a text-to-image model. These generated images are then used as image references in our \textbf{Dual-FDM} framework, where we disentangle the foreground using (${W}_k^{im}$) and the background using ($\overline{W}_k^{im}$). For quantitative evaluation, we compute the similarity between the disentangled images and the text prompts for foreground or background in Tab.~\ref{table:intuitions} using the CLIP model. In addition, we generate images from text prompts for the foreground or background using a text-to-image model, and then measure the similarity between the disentangled images and these generated images. Both CLIP and DINO models are employed to assess the image-level similarity, providing a comprehensive evaluation of the consistency of the foreground–background disentanglement. As illustrated in the left of Tab.~\ref{table:similar}, ${W}_k^{im}$ is able to disentangle more foreground information than $\overline{W}_k^{im}$, as evidenced by the higher scores in CLIP-T, CLIP-I, and DINO-I, whereas $\overline{W}_k^{im}$ is more effective at disentangling background information, as shown in the right of Tab.~\ref{table:similar}.

We also conduct additional ablation studies to investigate different strategies for suppressing foreground components and enhancing background components in the singular values. Specifically, we consider the hard suppression approach $\sigma_{\text{top-}n\%=0}$, which denotes setting the top $n\%$ of singular values to zero, and the soft suppression approach $\sigma e^{-\sigma}$, which represents a reweighting scheme that attenuates larger singular values while preserving smaller ones. Tab.~\ref{tab:singular} reports the results for all cases: overall, all the adjustment strategies show no significant difference in their ability to extract background information (measured by  CLIP-T and CLIP-I mean scores); the main variations are observed in the fluctuations across different images (measured by  CLIP-T and CLIP-I std scores). Our method demonstrates stable background disentanglement across diverse images. Setting different proportions of the top 10--50\% of singular values to zero does not yield a clear improvement in performance. This is because the hard partitioning of foreground and background components makes it difficult to adaptively determine the proportion of foreground and background for each  image, leading to limited performance gains and high variability across images.

Similarly, the soft suppression approach $\sigma e^{-\sigma}$ fails to effectively disentangle background for each image. While it adaptively suppresses foreground components, it does not explicitly enhance background components. As a result, the contribution of background components in the cross-attention remains limited, leading to incomplete background disentanglement. In contrast, our reciprocal operation $1/\sigma$ not only weakens the dominant foreground components but also actively amplifies the smaller singular values corresponding to background. This balanced adjustment ensures both suppression of foreground and enhancement of background, which results in more stable and consistent background disentanglement across images.

{We further analysis the singular values in the different cross-attention layers based on the singular value decomposition (SVD) on $W_{k}^{im}$ (Eq.(\ref{eq:cor}) of the mainbody) and $\overline{W}_{k}^{im}$ (Eq.(\ref{eq:style}) of the mainbody). For the singular value matrices obtained from different layers, we compute the proportion of the top n\% singular values relative to the sum of all singular values, named $\sum \sigma_{\text{top-n\%}}$. We also take the reciprocal of each singular value and perform the same calculation. In both cases, the proportions are evaluated for $n \in\{2, 5, 10, 20\}$. Tab.~\ref{tab:layer} indicates that: under the same spatial resolution, the proportions of the top n\% singular values exhibit highly consistent patterns across different layers. This consistency supports our claim that \textit{larger $\sigma_i$ tends to correspond to foreground-dominant components, whereas smaller $\sigma_i$ is more closely associated with background information} (Sec.~\ref{sec:back} of the mainbody), which confirms universally across layers. Furthermore, after inverting the singular values, the resulting distributions across layers also display similar behaviors, indicating that our style-disentangled matrices $\overline{W}_{k}^{im}$ effectively amplify the relatively weak background components.}
\begin{table*}[t]
\centering
\footnotesize
\caption{The examples of text prompts  for both foreground and background, foreground, and background to discussion intuitions for the rationale between ${W}_k^{im}$ and $\overline{W}_k^{im}$.}
\label{table:intuitions}
\begin{tabular}{c|c|c|c}
\toprule[1.5pt]
ID & Both foreground and background & Foreground & Background \\
\midrule
1 & A snow leopard is leaping across a fractured frozen expanse.
  & a snow leopard
  & the fractured frozen expanse \\

2 & A giant squid is drifting in a radiant deep chasm.
  & a giant squid
  & the radiant deep chasm \\

3 & A phoenix is glowing over an age-worn conflict field.
  & the phoenix
  & the age-worn conflict field \\

4 & A red-crowned crane is gliding through a crimson-lit horizon.
  & a red-crowned crane
  & the crimson horizon \\

5 & A polar bear is sitting beside a fractured icy dome-like form.
  & a polar bear
  & the fractured icy dome \\

6 & A chameleon is blending into a glare-soaked luminous corridor.
  & a chameleon
  & the luminous corridor \\

7 & A mountain gorilla is climbing an overrun observational platform.
  & a mountain gorilla
  & the overrun platform \\

8 & A humpback whale is singing beneath drifting frozen masses.
  & a humpback whale
  & the drifting frozen masses \\

9 & A komodo dragon is sunbathing on dark volcanic granules.
  & a komodo dragon
  & the volcanic granules \\

10 & A peregrine falcon is landing on a ruined high outpost.
   & a peregrine falcon
   & the ruined high outpost \\
\bottomrule[1.5pt]
\end{tabular}
\end{table*}

\begin{table*}[t]
\centering
\footnotesize
\caption{Quantitative comparisons of foreground–background disentanglement across different images via ${W}_k^{im}$ and $\overline{W}_k^{im}$. }
\label{table:similar}
\begin{tabular}{c|cccccc}
\toprule[1.5pt]
\multirow{2}{*}{Method}
& \multicolumn{3}{c}{Foreground}
& \multicolumn{3}{c}{Background} \\
\cmidrule(lr){2-4} \cmidrule(lr){5-7}
& CLIP-T$\uparrow$ &  CLIP-I$\uparrow$ & DINO-I$\uparrow$
& CLIP-T$\uparrow$ &  CLIP-I$\uparrow$ & DINO-I$\uparrow$ \\\midrule
${W}_k^{im}$      & \cellcolor{red!20}0.3115 & \cellcolor{red!20}0.8228 & \cellcolor{red!20}0.5774 & 0.2303 & 0.6157 & 0.1353 \\\midrule
$\overline{W}_k^{im}$ & 0.2629 & 0.7090 & 0.2591 & \cellcolor{red!20}0.2486 & \cellcolor{red!20}0.6421 & \cellcolor{red!20}0.1966 \\
\bottomrule[1.5pt]
\end{tabular}
\end{table*}

\begin{table*}[t]
\centering
\footnotesize
\caption{Ablation studies are conducted to evaluate three strategies for manipulating singular values: $\sigma_{\text{top-}n\%=0}$, which removes the largest $n\%$ of singular values by setting them to zero; $\sigma e^{-\sigma}$, a reweighting scheme that suppresses larger singular values while retaining smaller ones; and ours ($\frac{1}{\sigma}$), which inversely scales singular values. These strategies are examined to assess their effectiveness in suppressing foreground components and enhancing background components.}\label{tab:singular}
\begin{tabular}{c|cccc}
\toprule[1.5pt]
Method
& CLIP-T-Mean$\uparrow$
& CLIP-T-Std$\downarrow$
& CLIP-I-Mean$\uparrow$
& CLIP-I-Std$\downarrow$ \\
\midrule
$\sigma_{\text{top-}10\%=0}$
& 0.2469 (-0.68\%) & 0.0352 (+13.5\%)
&  \cellcolor{red!20}0.6450 (+0.45\%) & 0.0867 (+8.36\%) \\
$\sigma_{\text{top-}20\%=0}$
& 0.2496 (+0.40\%) & 0.0344 (+11.0\%)
& 0.6433 (+0.19\%) & 0.0806 (+0.62\%) \\
$\sigma_{\text{top-}30\%=0}$
&  \cellcolor{red!20}0.2500 (+0.56\%) & 0.0332 (+7.10\%)
& 0.6401 (-0.31\%) & 0.0842 (+5.11\%) \\
$\sigma_{\text{top-}40\%=0}$
& 0.2495 (+0.36\%) & 0.0334 (+7.74\%)
& 0.6382 (-0.30\%) & 0.0826 (+3.11\%) \\
$\sigma_{\text{top-}50\%=0}$
& 0.2499 (+0.52\%) & 0.0322 (+3.87\%)
& 0.6416 (-0.08\%) & 0.0868 (+8.36\%) \\\midrule
$\sigma e^{-\sigma}$
& 0.2498 (+0.49\%) & 0.0330 (+6.45\%)
& 0.6401 (-0.31\%) & 0.0883 (+10.2\%) \\\midrule
Ours ($\frac{1}{\sigma}$)
& 0.2486 &  \cellcolor{red!20}0.0310 & 0.6421 & \cellcolor{red!20} 0.0801 \\
\bottomrule[1.5pt]
\end{tabular}
\end{table*}

\begin{table*}[t]
\centering
\small
\caption{Summation of the top-$k\%$ singular values and their reciprocal counterparts across different cross-attention layers.}\label{tab:layer}
\begin{adjustbox}{max width=\textwidth}
\begin{tabular}{c c c c c c c c c c}
\toprule
\textbf{Layer Name} & \textbf{Res.} &
$\sum \sigma_{\text{top-2\%}}$ &
$\sum \sigma_{\text{top-5\%}}$ &
$\sum \sigma_{\text{top-10\%}}$ &
$\sum \sigma_{\text{top-20\%}}$ &
$\sum \frac{1}{\sigma}_{\text{top-2\%}}$ &
$\sum \frac{1}{\sigma}_{\text{top-5\%}}$ &
$\sum \frac{1}{\sigma}_{\text{top-10\%}}$ &
$\sum \frac{1}{\sigma}_{\text{top-20\%}}$ \\
\midrule
down\_blocks.1.0.0 & 64$\times$64 & 0.127 & 0.259 & 0.408 & 0.606 & 0.520 & 0.724 & 0.849 & 0.934 \\
down\_blocks.1.1.1 & 64$\times$64 & 0.147 & 0.279 & 0.431 & 0.635 & 0.562 & 0.756 & 0.869 & 0.944 \\
down\_blocks.2.0.0 & 32$\times$32 & 0.197 & 0.360 & 0.544 & 0.763 & 0.525 & 0.715 & 0.838 & 0.927 \\
down\_blocks.2.1.9 & 32$\times$32 & 0.228 & 0.392 & 0.562 & 0.751 & 0.552 & 0.733 & 0.849 & 0.932 \\
mid\_block.0.0      & 32$\times$32 & 0.208 & 0.384 & 0.573 & 0.790 & 0.533 & 0.721 & 0.842 & 0.929 \\
mid\_block.0.9      & 32$\times$32 & 0.234 & 0.381 & 0.536 & 0.719 & 0.563 & 0.739 & 0.853 & 0.934 \\
up\_blocks.0.0.0    & 32$\times$32 & 0.179 & 0.344 & 0.532 & 0.755 & 0.518 & 0.710 & 0.835 & 0.926 \\
up\_blocks.0.2.9    & 32$\times$32 & 0.285 & 0.489 & 0.652 & 0.783 & 0.556 & 0.735 & 0.850 & 0.933 \\
up\_blocks.1.0.0    & 64$\times$64 & 0.130 & 0.275 & 0.437 & 0.641 & 0.537 & 0.739 & 0.859 & 0.939 \\
up\_blocks.1.2.1    & 64$\times$64 & 0.150 & 0.292 & 0.448 & 0.647 & 0.527 & 0.733 & 0.856 & 0.938 \\
\bottomrule
\end{tabular}
\end{adjustbox}
\end{table*}

\section{The algorithm of customization style transfer and color style transfer}\label{sec1}
Due to the page limitation, we summarize the process of customization style transfer in Algorithm~\ref{algorithm1} and  color style transfer in Algorithm~\ref{algorithm2}.
To tackle the text-alignment issue in personalized image generation, in this paper, we study personalized generation based on dual references - customization and color and style reference - and propose a paradigm to disentangle these \textbf{\underline{Dual}} image references within \textbf{\underline{F}}requency-aware \textbf{\underline{D}}iffusion \textbf{\underline{M}}odels, dubbed \textbf{Dual-FDM}, for two classical personalized image generation tasks in a row: customization style transfer and color style transfer, by disentangling different frequency bands via mask strategy within frequency domain. For customization style transfer,  we replace the mid-frequency band of the background in the style reference with that from the foreground of the customized reference. For color style transfer, we substitute the low-frequency band of the background in the style reference with that from both the foreground and background of the color reference. Both the substituted frequency bands are used as the key and value to reconstruct the query foreground and background of the denoised personalized image.
For each task, we further compute the ratio of information entropy of the substituted frequency bands to adaptively modulate the denoising timesteps between the early and late stages.

\begin{algorithm*}[t]
	\caption{\textbf{Dual-FDM}$_{cust}$: Customization Style Transfer}
    \label{algorithm1}
	\LinesNumbered 
    \KwIn{the random Gaussian noise $z_{T}$, the text prompts $C_{te}$, the customized reference $C_{cu}$,  the style reference $C_{st}$, the denoising network $\epsilon_{\theta}$, the mid-frequency masks in the early denoising stages $m^{e}_{mid}$, the mid-frequency masks in the late denoising stages $m^{l}_{mid}$, the timestep $T$ and the decoder $\mathcal{D}$}
	\KwOut{The personalized generated image $I_{cust}$}
               Calculating $K^{m^{e}_{mid}}_{im_{cust}}$  and $V^{m^{e}_{mid}}_{im_{cust}}$ conditioned on $m^{e}_{mid}$ via Eq.(\textcolor[rgb]{0.00,0.00,1.00}{6}) of the mainbody for the early stage;

               Calculating $K^{m^{l}_{mid}}_{im_{cust}}$  and $V^{m^{l}_{mid}}_{im_{cust}}$ conditioned on $m^{l}_{mid}$ via Eq.(\textcolor[rgb]{0.00,0.00,1.00}{6}) of the mainbody for the late stage;

               Calculating the threshold $\Delta_{cust}$  conditioned on $K^{m^{e}_{mid}}_{im_{cust}}$ and $K^{m^{l}_{mid}}_{im_{cust}}$  via Eq.(\textcolor[rgb]{0.00,0.00,1.00}{11}) of the mainbody;

        \For{t = T, \ldots , $\frac{1}{1+\Delta_{cust}}$T}{
            Calculating $z_{t-1}$ from $z_{t}$ is performed in the decoupled cross-attention (Eq.(\textcolor[rgb]{0.00,0.00,1.00}{1}) and Eq.(\textcolor[rgb]{0.00,0.00,1.00}{5}) of the mainbody) of $\epsilon_{\theta} (C_{te}, C_{cu}, C_{st})$, the substituted frequency bands are used as the key $K^{m^{e}_{mid}}_{im_{cust}}$ and value $V^{m^{e}_{mid}}_{im_{cust}}$ to reconstruct the query foreground and background of the denoised personalized image;
    	}

        \For{t = $\frac{1}{1+\Delta_{cust}}$T, \ldots , 0}{
            Calculating $z_{t-1}$ from $z_{t}$ is performed in the decoupled cross-attention (Eq.(\textcolor[rgb]{0.00,0.00,1.00}{1}) and Eq.(\textcolor[rgb]{0.00,0.00,1.00}{5}) of the mainbody) of $\epsilon_{\theta} (C_{te}, C_{cu}, C_{st})$, the substituted frequency bands are used as the key $K^{m^{l}_{mid}}_{im_{cust}}$ and value $V^{m^{l}_{mid}}_{im_{cust}}$ to reconstruct the query foreground and background of the denoised personalized image;
    	}
 \Return Output the personalized generated image $I_{cust} = \mathcal{D}(z_{0})$
\end{algorithm*}

\begin{algorithm*}[t]
	\caption{\textbf{Dual-FDM}$_{cost}$: Color Style Transfer}
    \label{algorithm2}
	\LinesNumbered 
    \KwIn{the random Gaussian noise $z_{T}$, the text prompts $C_{te}$, the color reference $C_{co}$,  the style reference $C_{st}$, the denoising network $\epsilon_{\theta}$, the low-frequency masks in the early denoising stages $m^{e}_{low}$, the low-frequency masks in the late denoising stages $m^{l}_{low}$, the timestep $T$ and the decoder $\mathcal{D}$}
	\KwOut{The personalized generated image $I_{cost}$}
               Calculating $K^{m^{e}_{low}}_{im_{cost}}$  and $V^{m^{e}_{low}}_{im_{cost}}$ conditioned on $m^{e}_{low}$ via Eq.(\textcolor[rgb]{0.00,0.00,1.00}{9}) of the mainbody for the early stage;

               Calculating $K^{m^{l}_{low}}_{im_{cost}}$  and $V^{m^{l}_{low}}_{im_{cost}}$ conditioned on $m^{l}_{low}$ via Eq.(\textcolor[rgb]{0.00,0.00,1.00}{9}) of the mainbody for the late stage;

               Calculating the threshold $\Delta_{cost}$  conditioned on $K^{m^{e}_{low}}_{im_{cost}}$ and $K^{m^{l}_{low}}_{im_{cost}}$  via Eq.(\textcolor[rgb]{0.00,0.00,1.00}{11}) of the mainbody;

        \For{t = T, \ldots , $\frac{1}{1+\Delta_{cost}}$T}{
            Calculating $z_{t-1}$ from $z_{t}$ is performed in the decoupled cross-attention (Eq.(\textcolor[rgb]{0.00,0.00,1.00}{1}) and Eq.(\textcolor[rgb]{0.00,0.00,1.00}{8}) of the mainbody) of $\epsilon_{\theta} (C_{te}, C_{co}, C_{st})$, the substituted frequency bands are used as the key $K^{m^{e}_{low}}_{im_{cost}}$ and value $V^{m^{e}_{low}}_{im_{cost}}$ to reconstruct the query foreground and background of the denoised personalized image;
    	}

        \For{t = $\frac{1}{1+\Delta_{cost}}$T, \ldots , 0}{
            Calculating $z_{t-1}$ from $z_{t}$ is performed in the decoupled cross-attention (Eq.(\textcolor[rgb]{0.00,0.00,1.00}{1}) and Eq.(\textcolor[rgb]{0.00,0.00,1.00}{8}) of the mainbody) of $\epsilon_{\theta} (C_{te}, C_{co}, C_{st})$, the substituted frequency bands are used as the key $K^{m^{l}_{low}}_{im_{cost}}$ and value $V^{m^{l}_{low}}_{im_{cost}}$ to reconstruct the query foreground and background of the denoised personalized image;
    	}
 \Return Output the personalized generated image $I_{cost} = \mathcal{D}(z_{0})$
\end{algorithm*}

\section{More details about the implementation details}\label{sec3}
As mentioned in Sec.3.1 of the mainbody, due to page limitation, we further provide more details about the implementation details as follows:

\textbf{Datasets.} We evaluate \textbf{Dual-FDM} on combined datasets covering image customization, style transfer, and text-to-image (T2I) generation:
\begin{itemize}
\item \textbf{Image Customization}: 1) \textbf{DreamBench} \cite{ruiz2023dreambooth} + \textbf{TI benchmark} \cite{gal2022image}: a well-known single-object personalized image generation dataset comprising 38 subjects, covering unique objects and pets such as backpacks, stuffed animals, dogs, cats, sunglasses, and cartoons, with each subject associated with 25 prompts;  2) \textbf{DreamBench++} \cite{pengdreambench++}: the latest human-aligned personalized image generation benchmark, comprising 65 object classes, 20 human classes, and 45 animal classes. For each image, 9 text prompts are generated by GPT-4o to span a range of difficulty levels.
\item \textbf{Style Transfer}:  1) \textbf{IS-Plus benchmark} \cite{wang2024instantstyle1}: a style evaluation benchmark containing 104 images across diverse styles; 2) \textbf{StyleBench} \cite{gao2024styleshot}: a style evaluation benchmark covering 73 distinct styles, ranging from paintings, flat illustrations, and 3D rendering to sculptures with varying materials. For each style, 5–7 distinct images with variations are collected. In total, StyleBench contains 490 images across diverse styles.
\item \textbf{T2I generation}: \textbf{T2I-CompBench++} \cite{huang2025t2icompbench++}: an enhanced benchmark for compositional text-to-image generation. T2I-CompBench++ comprises 8,000 compositional text prompts categorized into four primary groups: attribute binding, object relationships, generative numeracy, and complex compositions. These are further divided into eight sub-categories.
\end{itemize}
Based on the above, we construct several combined datasets for different joint tasks: 1) for customization  style transfer (small-scale), we concatenate \textbf{DreamBench} and \textbf{TI benchmark} as the customization references, and use \textbf{IS-Plus benchmark} as the style references; 2) for customization  style transfer (large-scale), we take \textbf{DreamBench++} as the customization references and \textbf{StyleBench} as the style references; 3) for color style transfer, we combine \textbf{InstantStyle-Plus} and \textbf{StyleBench} as the color and style references, while text prompts are provided by \textbf{T2I-CompBench++} for stylized image generation.

\textbf{Metrics.} We consider 9 evaluation metrics from three aspects to evaluate our \textbf{Dual-FDM}:
\begin{itemize}
\item \textbf{Text and Image Reference Alignment}: we follow previous methods and adopt two representative models, \textbf{CLIP}\footnote{Since CLIP is trained on real image–text (T) pairs, it is particularly sensitive to image style, which may distort text–image similarity. To alleviate this, we propose CLIP-TS by measuring similarity with style reference (S), thereby reducing the style influence . In addition, the generated background aligned with text prompts can interfere with the customized reference (C); hence we propose CLIP-TC, which incorporates text prompts to suppress such interference. We also provide CLIP-T (image–text similarity) and CLIP-CS (image–image reference similarity) for comparison.} and \textbf{DINO} \cite{caron2021emerging}, for evaluation. Specifically, \textbf{CLIP-T} measures the similarity between generated images and the given text prompts; \textbf{CLIP-TC} and \textbf{CLIP-TS} evaluate the similarity between generated images and customized (or style) reference images while jointly considering the text prompts; finally, \textbf{CLIP-CS} assesses the similarity between generated images and dual-reference images.
\item \textbf{Image Generation Quality}: we employ Image Reward (\textbf{IR}) \cite{xu2023imagereward} to model human preferences for image quality and \textbf{HPS V2} \cite{wu2023human} that combines perceptual studies with deep learning for comprehensive visual and semantic assessment. Both metrics measure the semantics consistency of the high-quality image generation aligned with human standards;
\item \textbf{Time Complexity}: We employ the inference time (\textbf{Times}) and Video Random Access Memory (\textbf{VRAM}) consumption to evaluate the computational efficiency and memory footprint of different methods during image generation.
\end{itemize}

\textbf{Model Configuration.} Our main experiments are conducted on the pre-trained IP-Adapter-Plus \cite{ye2023ip} with SDXL \cite{podell2023sdxl} model as the text-to-image diffusion model, configured with 30 sampling steps and and a fixed random seed of 42, to ensure high-quality outputs. We use the recommended hyper-parameters for each method across all images to ensure fair comparison. All experiments are implemented in PyTorch and conducted on a  NVIDIA RTX 4090 GPU.

\section{More Analysis of Computational Efficiency in the Sec.3.2 of the Mainbody}\label{sec4_0}
As indiacted in the mainbody, we further provide the inference time and VRAM consumption of our proposed \textbf{Dual-FDM$_{cust}$} model, and compared it with several representative baselines, which are reported in Table.~\ref{tab:performance_time}. The inference time of our \textbf{Dual-FDM$_{cust}$} is 10.26s. This demonstrates superior efficiency compared with state-of-the-art methods. Our \textbf{Dual-FDM$_{cust}$} also achieves the the lowest result in VRAM consumption. Specifically, \textbf{SSR-Encoder} adopts SD v1.5 and \textbf{Dreamcache} is designed for lightweight usage, so both consume less VRAM.

\begin{table}[t]
\setlength{\abovecaptionskip}{-2pt}
\setlength{\belowcaptionskip}{0pt}
\caption{Efficiency comparisons between \textbf{Dual-FDM$_{cust}$} and other diffusion-based personalization models.}
\label{tab:performance_time}
\centering
\scriptsize
\renewcommand{\arraystretch}{0.9}
\setlength{\tabcolsep}{0.5pt}
\begin{adjustbox}{max width=\linewidth}
\begin{tabular}{@{}l *{7}{c}@{}}
\toprule[1.5pt]
& \makecell[c]{\textbf{IP-Adapter} \\ Arxiv'23} &
  \makecell[c]{\textbf{SSR-Encoder} \\ CVPR'24} &
  \makecell[c]{\textbf{MIP-Adapter} \\ AAAI'25} &
  \makecell[c]{\textbf{Dreamcache} \\ CVPR'25} &
  \makecell[c]{\textbf{PatchDPO} \\ CVPR'25} &
  \makecell[c]{\textbf{MS-Diffusion} \\ ICLR'25} &
  {\textbf{Dual-FDM$_{cust}$}}\\
\midrule
\multicolumn{8}{c}{\textsc{\{Baseline\} + StyleID (CVPR'24)}} \\
\midrule
Time(s) $\downarrow$ & 13.30 & \cellcolor{blue!20}11.03 & 15.85 & 12.17 & 13.11 & 14.68 &\cellcolor{red!20}10.26 \\
VRAM(GB) $\downarrow$ & 16.92 & 13.88 & 16.82 & \cellcolor{blue!20}13.80 & 16.92 & 16.85 &\cellcolor{red!20}12.48 \\
\bottomrule[1.5pt]
\end{tabular}
\end{adjustbox}
\vspace{-1.5em}
\end{table}

\begin{figure*}[t]
  \centering
  \includegraphics[width=\textwidth]{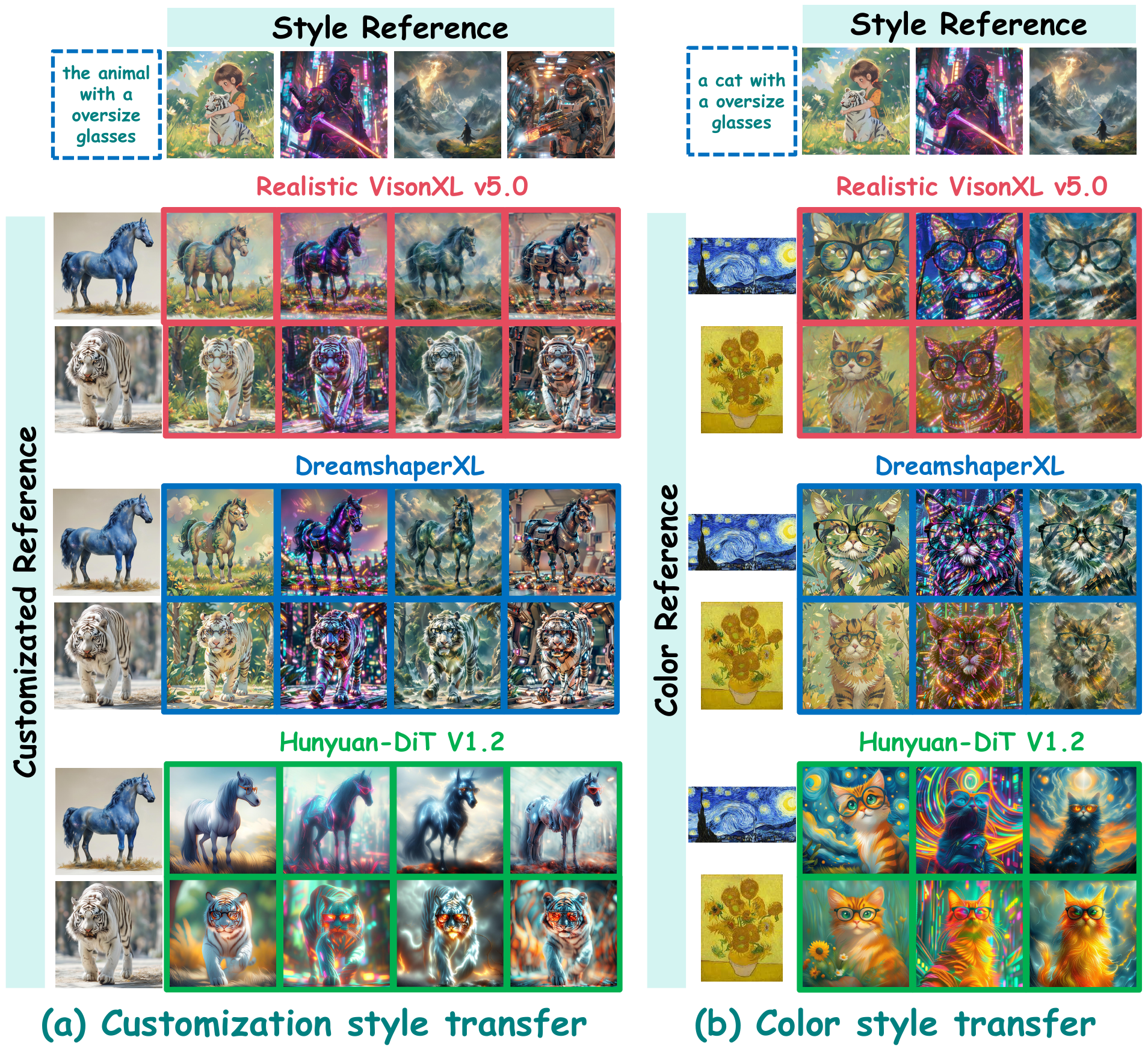}
  \caption{Generation results of different T2I models.}
  \label{fig:ab_visual3}
\end{figure*}

\section{More generation results of different T2I Models}\label{sec4}
As mentioned in Sec.3.2 of the mainbody, we further report the generated results of several representative text-to-image (T2I) models in Fig.~\ref{fig:ab_visual3}, including different finetuned variants of Stable Diffusion XL (SDXL), namely RealisticVisionXL and DreamshaperXL, as well as the DiT-based approach HunyuanDiT. These three models exhibit distinct characteristics due to differences in their training datasets:
\begin{itemize}[leftmargin=*, topsep=0pt, parsep=0pt]
    \item \textbf{RealisticVisionXL}: fine-tuned on high-quality, photography-oriented datasets, emphasizing realism, lighting, and fine-grained textures, making it particularly effective for photorealistic inpainting tasks. As shown in Fig.~\ref{fig:ab_visual3}(a), the customized contents and the style can be well balanced.
    \item \textbf{DreamshaperXL}: incorporates a mixture of artistic and stylized datasets, leading to visually rich outputs with enhanced creativity and aesthetic variations, though sometimes at the expense of strict semantic fidelity to the input prompts. As shown in Fig.~\ref{fig:ab_visual3}, the personalized images emphasize style more strongly.
    \item \textbf{HunyuanDiT}: trained on large-scale, multi-domain datasets covering both natural scenes and human-centric images, which provides stronger generalization ability across diverse prompts and better semantic alignment. Benefiting from its realism-oriented training, it is more favorable to the image customization of our \textbf{Dual-FDM}, as illustrated in  Fig.~\ref{fig:ab_visual3}(a).
\end{itemize}

\begin{figure*}[t]
  \centering
  \includegraphics[width=\textwidth]{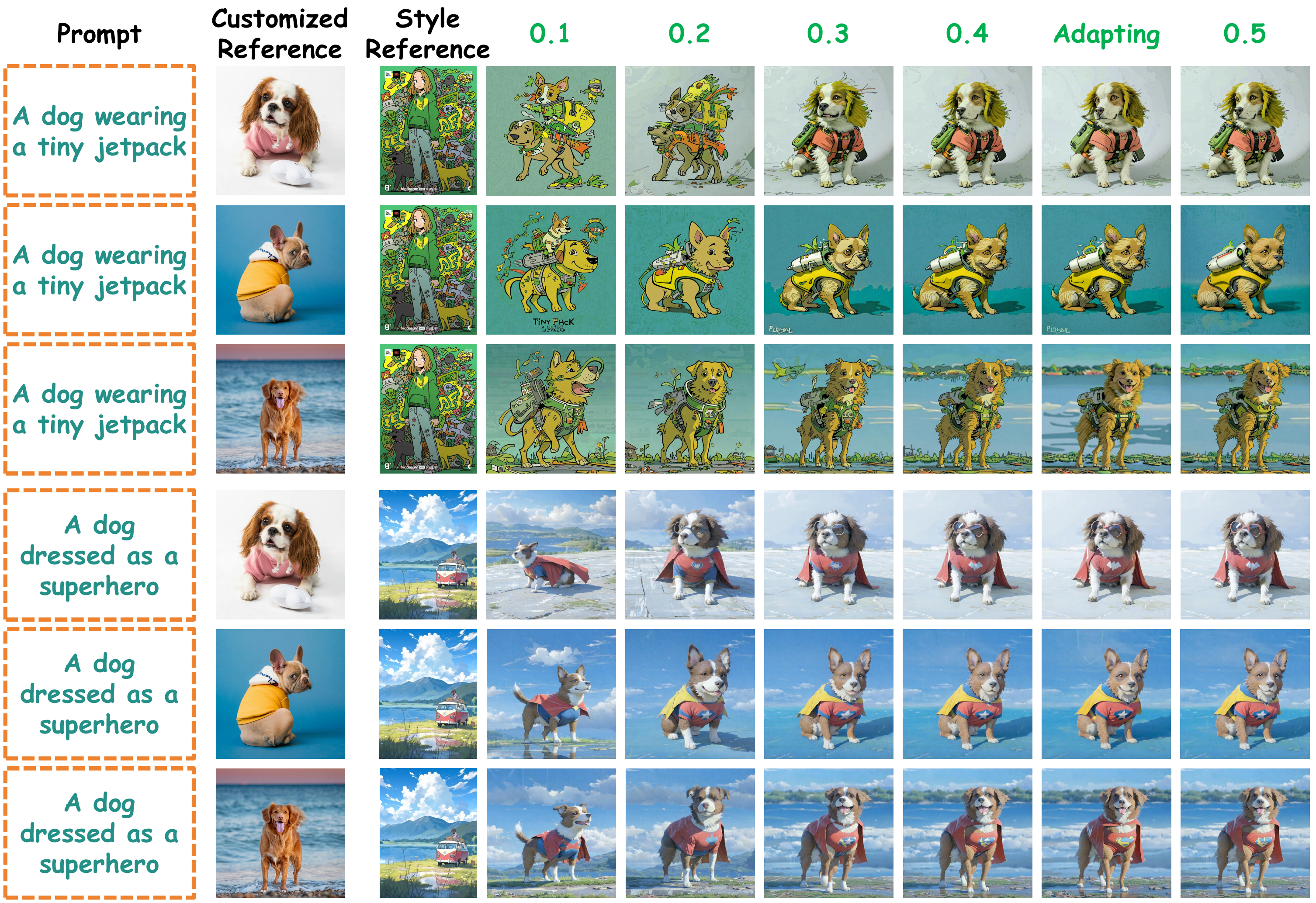}
  \caption{Ablation studies about the influence of different denoising timesteps between the early and late stages on generating the personalized images.}
  \label{fig:ab_visual1}
\end{figure*}

\section{Additional generation results for the ablation study about the denoising timesteps between the early and late stages}\label{sec5}
As mentioned in  Sec.3.3 of the mainbody, due to page limitation, we further provide more generation results to analyse the influence of the denoising timesteps between the early and late stages on generating the personalized images. Fig.~\ref{fig:ab_visual1} reports that when the proportion is set to 0.1, the generated images show the highest similarity to the style reference while being least similar to the customized reference; conversely, at 0.5, the opposite trend is observed. We can also conclude that our \textbf{Adapting} (Eq.(11) of the mainbody) lies between the results of the 0.4 and 0.5 proportions, demonstrating that our dynamic adjustment strategy can effectively regulate the ratio of early to late stages, leading to a well-balanced performance in terms of both text alignment and dual-image similarity, confirming the intuition of Sec.2.5 of the mainbody.

\section{Additional generation results for the ablation study about the different masks to filter mid-and-low frequency bands}\label{sec6}
As mentioned in  Sec.3.3 of the mainbody, due to page limitation, we further provide more generation results to analyse the the different masks to filter mid-and-low frequency bands on generating the personalized images.  For the mid-frequency band, we define different thresholds $(th_{mp1}, th_{mp2})$, where $(th_{mp1}=0.1, th_{mp2}=0.9)$ represents the widest mid-frequency range and $(th_{mp1}=0.4, th_{mp2}=0.6)$ represents a narrower range. Fig.~\ref{fig:ab_visual2m} suggests the following:  when the mid-frequency range is the widest (\textit{i.e.}, $th_{mp1}=0.1, th_{mp2}=0.9$), the generated images show the highest similarity to the {customized reference} and the lowest similarity to the {style reference}. As the mid-frequency range becomes narrower (\textit{i.e.}, $th_{mp1}=0.4, th_{mp2}=0.6$), the opposite trend is observed, suggesting that a narrower mid-frequency band enhances the representation of style features. For the low-frequency band, we vary $th_{lp}\in\{0.8,0.6,0.4,0.2\}$, where $th_{lp}=0.8$ indicates the largest proportion of low-frequency components. Decreasing $th_{lp}$ from 0.8 to 0.2 (thus reducing the proportion of low-frequency bands) leads to higher similarity between the generated images and the style reference, but worsens color consistency, as illustrated in Fig.~\ref{fig:ab_visual2l}, which is consistent with Sec.2.4 of the mainbody.

\begin{figure*}[t]
  \centering
  \includegraphics[width=\textwidth]{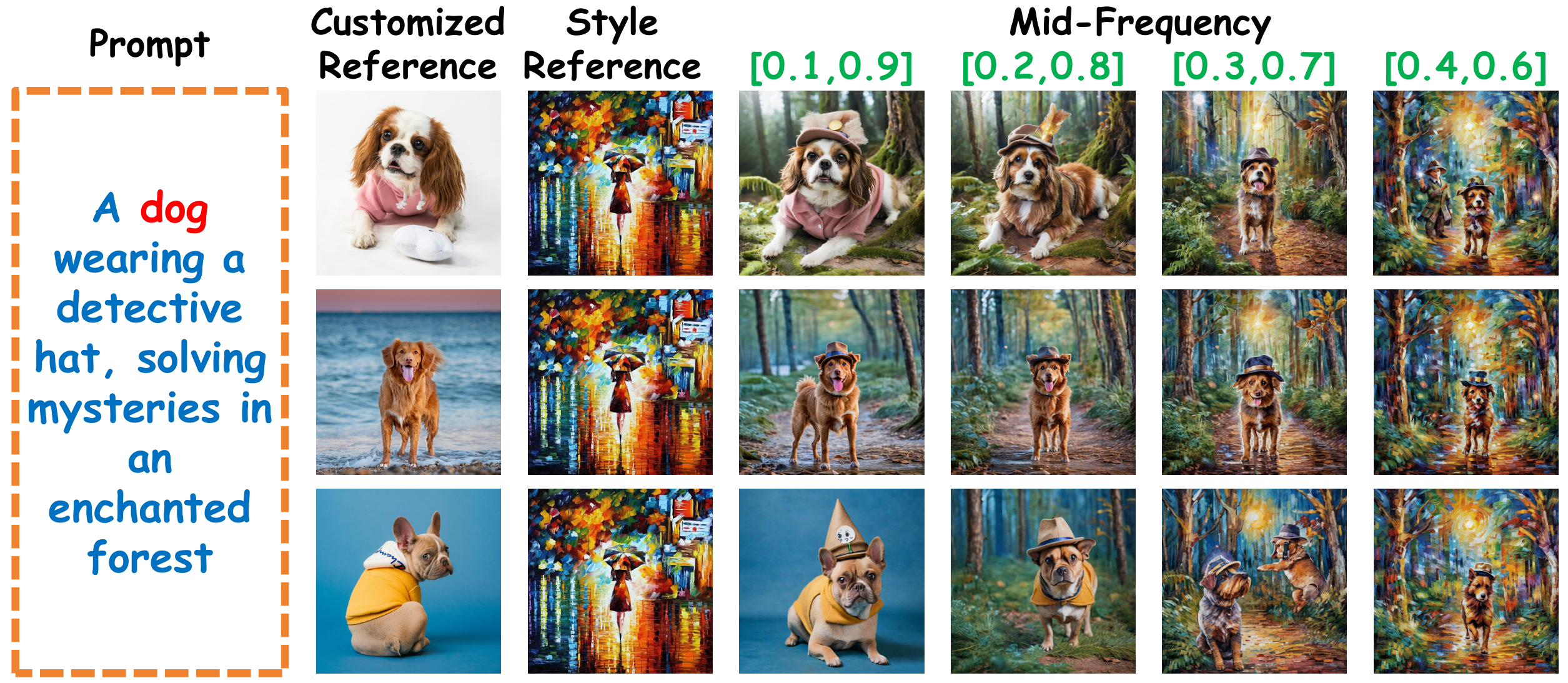}
  \caption{Ablation studies about the influence of the different masks to filter mid-frequency band on generating the personalized images.}
  \label{fig:ab_visual2m}
\end{figure*}

\begin{figure*}[t]
  \centering
  \includegraphics[width=\textwidth]{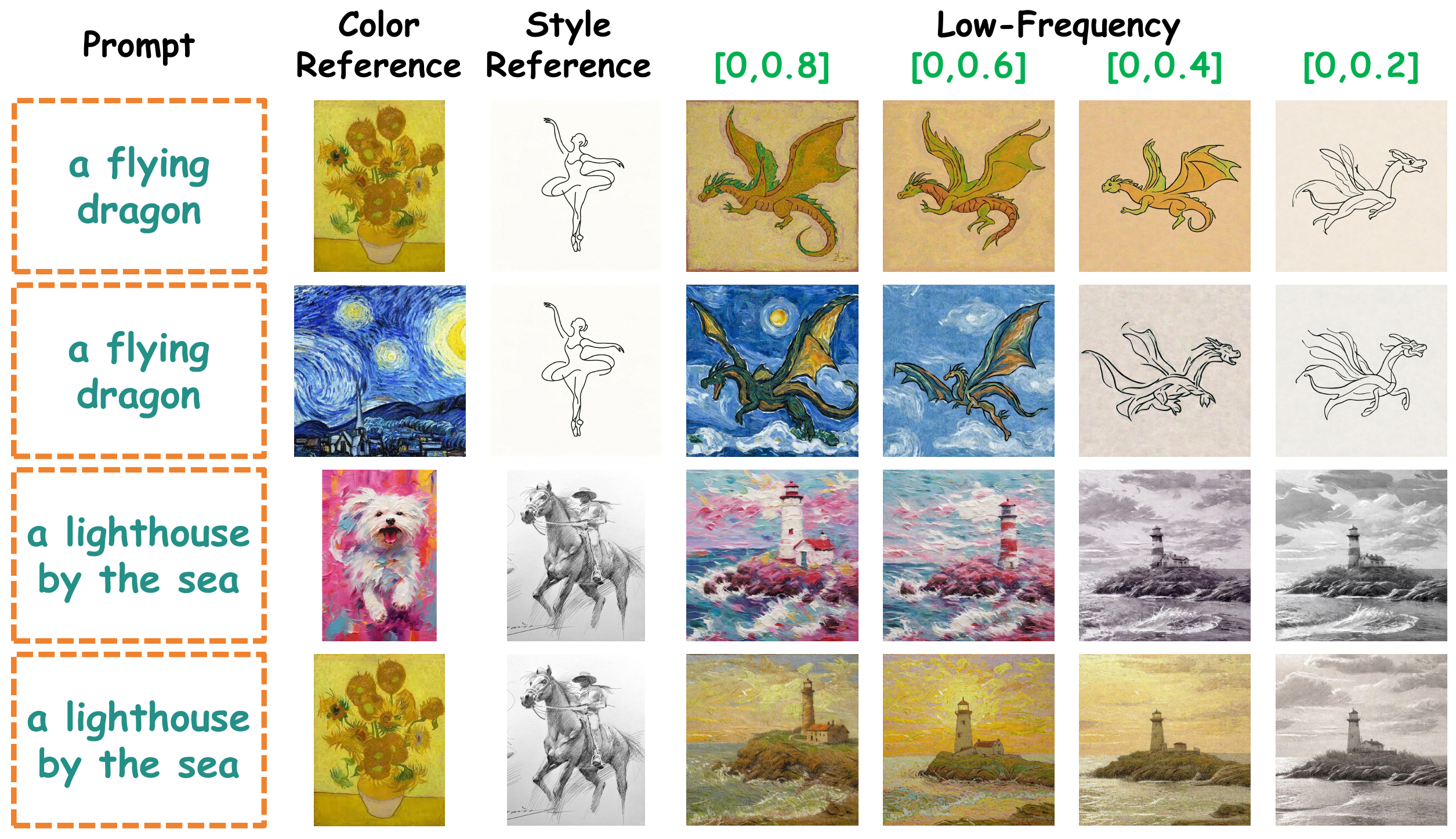}
  \caption{Ablation studies about the influence of the different masks to filter low-frequency band on generating the personalized images.}
  \label{fig:ab_visual2l}
\end{figure*}

\section{Limitations and Broader Impacts}\label{secli}

\noindent \textbf{Limitations.} As illustrated in Fig.~\ref{fig:Failure}, for customization style transfer, when the style reference contains extremely sparse patterns (\textit{e.g.}, line-based structures), the mid-frequency customized cues may \textit{{attenuate high-frequency style details, leading to incomplete style information}}. We plan to address this issue in our future work.

\noindent \textbf{Broader Impacts.} Our \textbf{Dual-FDM} can generate the personalized image according to the text prompt and reference images, our method may raise certain ethical concerns. The personalized image could potentially be misused in the creation of misleading information. We strongly advocate for the establishment of clear accountability in the use of such technologies, along with enhanced legal and technical oversight, to ensure they are applied responsibly.

\begin{figure*}[t]
  \centering
  \includegraphics[width=0.5\textwidth]{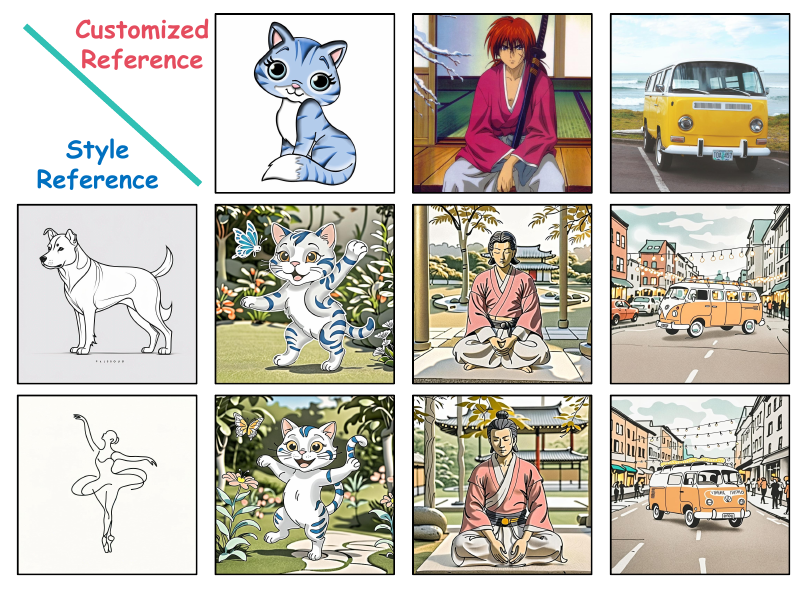}
  \caption{Failure examples for customization style transfer}
  \label{fig:Failure}
\end{figure*}

%

\section{Discussion on Additional Metrics and Style-Transfer Baselines}
\label{sec:style_transfer_baselines}

To provide a more comprehensive evaluation of style preservation in personalized image generation, we implemented additional style-transfer-related baselines, including {ZipLoRA \cite{shah2024ziplora}, B-LoRA \cite{frenkel2024implicit}, and UnZipLoRA \cite{liu2025unziplora}}. We further report the Contrastive Style Descriptor (CSD) metric and conduct a user study, as requested. For reference, we also include CLIP- and DINO-based metrics. The quantitative results are summarized in Table~\ref{tab:style_transfer_comparison}.

\begin{table}[t]
\centering
\caption{Quantitative comparison with style-transfer-related baselines. Higher values indicate better performance.}
\label{tab:style_transfer_comparison}
\resizebox{\linewidth}{!}{
\begin{tabular}{lccccc}
\hline
\textbf{Method} &
\textbf{CSD $\uparrow$} &
\textbf{CLIP-T $\uparrow$} &
\textbf{CLIP-I $\uparrow$} &
\textbf{DINO-I $\uparrow$} &
\textbf{User Preferred (\%) $\uparrow$} \\
\hline
ZipLoRA \cite{shah2024ziplora}  & 0.307 & 0.138 & 0.594 & 0.311 & 89\% \\
B-LoRA \cite{frenkel2024implicit}   & 0.363 & \textbf{0.173} & 0.666 & 0.353 & 72\% \\
UnZipLoRA \cite{liu2025unziplora}& 0.501 & 0.156 & 0.697 & 0.522 & 68\% \\
\textbf{Ours} &
\textbf{0.514} & 0.172 & \textbf{0.703} &
\textbf{0.559} & -- \\
\hline
\end{tabular}
}
\end{table}

As shown in Table~\ref{tab:style_transfer_comparison}, our method achieves the best CSD, CLIP-I, and DINO-I scores, while remaining competitive in terms of CLIP-T. In particular, CSD measures style similarity using a contrastively learned style representation, making it more suitable for evaluating fine-grained style preservation than general-purpose semantic metrics. These results demonstrate that our method more effectively preserves the reference style while maintaining strong semantic alignment and image-level consistency.

For the user study, 10 participants conducted pairwise comparisons between our method and each baseline and selected the result with better style preservation and overall generation quality. Our method was preferred over ZipLoRA, B-LoRA, and UnZipLoRA in 89\%, 72\%, and 68\% of the comparisons, respectively, further confirming its advantage in perceptual style preservation.

\section{More Results on Extreme Scenarios}\label{sec:extreme_scenarios}
We conduct additional experiments under extreme scenarios to investigate the effects of dense customized references and style references with varying levels of sparsity on the generated results. As shown in Table~\ref{tab:extreme_scenarios}, the results reveal a clear trade-off between style fidelity and customized-content preservation under extreme dual-reference settings. Specifically, dense customized references improve style fidelity, increasing CLIP-S by 18.39\%, but simultaneously reduce text alignment and content preservation, with CLIP-T and CLIP-C decreasing by 9.13\% and 16.81\%, respectively. Conversely, sparse settings achieve better alignment with the text prompt and higher similarity to the customized content, although at the cost of reduced style fidelity. These findings highlight the inherent challenge of balancing style transfer and customized-content preservation when handling extreme dual-reference scenarios.
\begin{table}[t]
\centering
\caption{Quantitative comparison under extreme dual-reference scenarios. Relative changes are computed from sparse to dense settings.}
\label{tab:extreme_scenarios}
\resizebox{0.8\linewidth}{!}{
\begin{tabular}{lccc}
\hline
\textbf{Setting} &
\textbf{CLIP-T $\uparrow$} &
\textbf{CLIP-C $\uparrow$} &
\textbf{CLIP-S $\uparrow$} \\
\hline
Sparse1 & 0.3074 & 0.8398 & 0.6445 \\
Sparse2 & 0.3159 & 0.8335 & 0.5586 \\
Dense1  & 0.2842 & 0.6865 & 0.7109 \\
Dense2  & 0.2822 & 0.7056 & 0.7134 \\
\hline
\textbf{Sparse Mean} &
\textbf{0.3117} & \textbf{0.8367} & \textbf{0.6016} \\
\textbf{Dense Mean} &
\textbf{0.2832} & \textbf{0.6961} & \textbf{0.7122} \\
\hline
\textbf{Relative Change (\%)} &
\textbf{-9.13\%} &
\textbf{-16.81\%} &
\textbf{+18.39\%} \\
\hline
\end{tabular}
}
\end{table}

\section{Related work}\label{rel} 

\textbf{Multi-condition control in diffusion models}

Controllable text-to-image diffusion models \cite{wang2024text,nair2024maxfusion,yang2025dc,cohen2025conditional} increasingly incorporate multiple visual, spatial, or semantic conditions in addition to text prompts. However, heterogeneous conditions may be imperfectly aligned and compete during denoising, causing one condition to dominate the generation or suppress text guidance. These methods are primarily designed for a sparse set of heterogeneous conditions, where the conditions often exhibit strong semantic or structural correlations. Independent of our model, they cannot be directly applied to dual-image-reference personalized generation task, which requires explicitly disentangling and independently controlling the different information provided by the two references.

\textbf{Style transfer}

Content–style disentanglement methods \cite{shah2024ziplora,frenkel2024implicit,liu2025unziplora} aim to preserve subject identity while transferring the visual appearance of a style reference. ZipLoRA \cite{shah2024ziplora} independently learns subject and style LoRAs and optimizes their merging coefficients to reduce interference between the two representations. B-LoRA \cite{frenkel2024implicit} achieves implicit content–style separation by learning LoRA parameters in selected layers of a pretrained diffusion model. UnZipLoRA \cite{liu2025unziplora} disentangles subject and style from a single stylized image by jointly learning two separate LoRAs with prompt- and layer-level separation. However, these approaches primarily perform disentanglement of a single image reference in the model-parameter space and generally require additional LoRA training or optimization. They do not explicitly exploit the frequency characteristics of the reference images.

\textbf{Frequency-aware methods in diffusion models}

Frequency-aware diffusion methods \cite{lee2025beta,issachar2026dype,ren2025fds,z2026colorful} exploit spectral properties for image generation and editing. These methods either operate without image references or rely on only one source or conditioning image. Therefore, they use frequency information for task-specific guidance rather than disentangling complementary attributes across multiple references. In contrast, our method decomposes and composes dual image references, selectively integrating subject, color, and texture information within a unified framework.

%% file: checklist.tex
\section*{NeurIPS Paper Checklist}

\begin{enumerate}

\item {\bf Claims}
    \item[] Question: Do the main claims made in the abstract and introduction accurately reflect the paper's contributions and scope?
    \item[] Answer:  \answerYes{} 
    \item[] Justification: We confirm the main claims reflect the paper’s contributions and scope.
    \item[] Guidelines:
    \begin{itemize}
        \item The answer \answerNA{} means that the abstract and introduction do not include the claims made in the paper.
        \item The abstract and/or introduction should clearly state the claims made, including the contributions made in the paper and important assumptions and limitations. A \answerNo{} or \answerNA{} answer to this question will not be perceived well by the reviewers. 
        \item The claims made should match theoretical and experimental results, and reflect how much the results can be expected to generalize to other settings. 
        \item It is fine to include aspirational goals as motivation as long as it is clear that these goals are not attained by the paper. 
    \end{itemize}

\item {\bf Limitations}
    \item[] Question: Does the paper discuss the limitations of the work performed by the authors?
    \item[] Answer: \answerYes{} 
    \item[] Justification: The paper have conducted a discussion of limiations in the Appendix.
    \item[] Guidelines:
    \begin{itemize}
        \item The answer \answerNA{} means that the paper has no limitation while the answer \answerNo{} means that the paper has limitations, but those are not discussed in the paper. 
        \item The authors are encouraged to create a separate ``Limitations'' section in their paper.
        \item The paper should point out any strong assumptions and how robust the results are to violations of these assumptions (e.g., independence assumptions, noiseless settings, model well-specification, asymptotic approximations only holding locally). The authors should reflect on how these assumptions might be violated in practice and what the implications would be.
        \item The authors should reflect on the scope of the claims made, e.g., if the approach was only tested on a few datasets or with a few runs. In general, empirical results often depend on implicit assumptions, which should be articulated.
        \item The authors should reflect on the factors that influence the performance of the approach. For example, a facial recognition algorithm may perform poorly when image resolution is low or images are taken in low lighting. Or a speech-to-text system might not be used reliably to provide closed captions for online lectures because it fails to handle technical jargon.
        \item The authors should discuss the computational efficiency of the proposed algorithms and how they scale with dataset size.
        \item If applicable, the authors should discuss possible limitations of their approach to address problems of privacy and fairness.
        \item While the authors might fear that complete honesty about limitations might be used by reviewers as grounds for rejection, a worse outcome might be that reviewers discover limitations that aren't acknowledged in the paper. The authors should use their best judgment and recognize that individual actions in favor of transparency play an important role in developing norms that preserve the integrity of the community. Reviewers will be specifically instructed to not penalize honesty concerning limitations.
    \end{itemize}

\item {\bf Theory assumptions and proofs}
    \item[] Question: For each theoretical result, does the paper provide the full set of assumptions and a complete (and correct) proof?
    \item[] Answer: \answerNA{} 
    \item[] Justification: We did not include theoretical results in our paper.
    \item[] Guidelines:
    \begin{itemize}
        \item The answer \answerNA{} means that the paper does not include theoretical results. 
        \item All the theorems, formulas, and proofs in the paper should be numbered and cross-referenced.
        \item All assumptions should be clearly stated or referenced in the statement of any theorems.
        \item The proofs can either appear in the main paper or the supplemental material, but if they appear in the supplemental material, the authors are encouraged to provide a short proof sketch to provide intuition. 
        \item Inversely, any informal proof provided in the core of the paper should be complemented by formal proofs provided in appendix or supplemental material.
        \item Theorems and Lemmas that the proof relies upon should be properly referenced. 
    \end{itemize}

    \item {\bf Experimental result reproducibility}
    \item[] Question: Does the paper fully disclose all the information needed to reproduce the main experimental results of the paper to the extent that it affects the main claims and/or conclusions of the paper (regardless of whether the code and data are provided or not)?
    \item[] Answer: \answerYes{} 
    \item[] Justification:  We have described the model details in the implementation details in Sec.~\ref{sec3} of the Appendix and carefully described the experimental metrics in the experimental chapter. The information we provide is sufficient and detailed for replication purposes.
    \item[] Guidelines:
    \begin{itemize}
        \item The answer \answerNA{} means that the paper does not include experiments.
        \item If the paper includes experiments, a \answerNo{} answer to this question will not be perceived well by the reviewers: Making the paper reproducible is important, regardless of whether the code and data are provided or not.
        \item If the contribution is a dataset and\slash or model, the authors should describe the steps taken to make their results reproducible or verifiable. 
        \item Depending on the contribution, reproducibility can be accomplished in various ways. For example, if the contribution is a novel architecture, describing the architecture fully might suffice, or if the contribution is a specific model and empirical evaluation, it may be necessary to either make it possible for others to replicate the model with the same dataset, or provide access to the model. In general. releasing code and data is often one good way to accomplish this, but reproducibility can also be provided via detailed instructions for how to replicate the results, access to a hosted model (e.g., in the case of a large language model), releasing of a model checkpoint, or other means that are appropriate to the research performed.
        \item While NeurIPS does not require releasing code, the conference does require all submissions to provide some reasonable avenue for reproducibility, which may depend on the nature of the contribution. For example
        \begin{enumerate}
            \item If the contribution is primarily a new algorithm, the paper should make it clear how to reproduce that algorithm.
            \item If the contribution is primarily a new model architecture, the paper should describe the architecture clearly and fully.
            \item If the contribution is a new model (e.g., a large language model), then there should either be a way to access this model for reproducing the results or a way to reproduce the model (e.g., with an open-source dataset or instructions for how to construct the dataset).
            \item We recognize that reproducibility may be tricky in some cases, in which case authors are welcome to describe the particular way they provide for reproducibility. In the case of closed-source models, it may be that access to the model is limited in some way (e.g., to registered users), but it should be possible for other researchers to have some path to reproducing or verifying the results.
        \end{enumerate}
    \end{itemize}

\item {\bf Open access to data and code}
    \item[] Question: Does the paper provide open access to the data and code, with sufficient instructions to faithfully reproduce the main experimental results, as described in supplemental material?
    \item[] Answer: \answerYes{} 
    \item[] Justification: We make a detailed description of the implementation details to ensure that they are repeatable and use publicly available personalized image generation datasets. We provide our code in the Appendix.
    \item[] Guidelines:
    \begin{itemize}
        \item The answer \answerNA{} means that paper does not include experiments requiring code.
        \item Please see the NeurIPS code and data submission guidelines (\url{https://neurips.cc/public/guides/CodeSubmissionPolicy}) for more details.
        \item While we encourage the release of code and data, we understand that this might not be possible, so \answerNo{} is an acceptable answer. Papers cannot be rejected simply for not including code, unless this is central to the contribution (e.g., for a new open-source benchmark).
        \item The instructions should contain the exact command and environment needed to run to reproduce the results. See the NeurIPS code and data submission guidelines (\url{https://neurips.cc/public/guides/CodeSubmissionPolicy}) for more details.
        \item The authors should provide instructions on data access and preparation, including how to access the raw data, preprocessed data, intermediate data, and generated data, etc.
        \item The authors should provide scripts to reproduce all experimental results for the new proposed method and baselines. If only a subset of experiments are reproducible, they should state which ones are omitted from the script and why.
        \item At submission time, to preserve anonymity, the authors should release anonymized versions (if applicable).
        \item Providing as much information as possible in supplemental material (appended to the paper) is recommended, but including URLs to data and code is permitted.
    \end{itemize}

\item {\bf Experimental setting/details}
    \item[] Question: Does the paper specify all the training and test details (e.g., data splits, hyperparameters, how they were chosen, type of optimizer) necessary to understand the results?
    \item[] Answer: \answerYes{} 
    \item[] Justification: We have carried out a detailed narration in the implementation details in Sec.~\ref{sec3} of the Appendix.
    \item[] Guidelines:
    \begin{itemize}
        \item The answer \answerNA{} means that the paper does not include experiments.
        \item The experimental setting should be presented in the core of the paper to a level of detail that is necessary to appreciate the results and make sense of them.
        \item The full details can be provided either with the code, in appendix, or as supplemental material.
    \end{itemize}

\item {\bf Experiment statistical significance}
    \item[] Question: Does the paper report error bars suitably and correctly defined or other appropriate information about the statistical significance of the experiments?
    \item[] Answer: \answerNA{}. 
    \item[] Justification: Based on our experimental experience, the reproducibility of the experiments involved in this work is high, with results that are replicable and stable, rather than simply reporting the highest outcomes. Additionally, previous related work \cite{ye2023ip,aiello2025dreamcache,huang2025resolving,zhang2024ssr,huang2025resolving} has also not reported error bars. We thus do not run the statistical significance test.
    \item[] Guidelines:
    \begin{itemize}
        \item The answer \answerNA{} means that the paper does not include experiments.
        \item The authors should answer \answerYes{} if the results are accompanied by error bars, confidence intervals, or statistical significance tests, at least for the experiments that support the main claims of the paper.
        \item The factors of variability that the error bars are capturing should be clearly stated (for example, train/test split, initialization, random drawing of some parameter, or overall run with given experimental conditions).
        \item The method for calculating the error bars should be explained (closed form formula, call to a library function, bootstrap, etc.)
        \item The assumptions made should be given (e.g., Normally distributed errors).
        \item It should be clear whether the error bar is the standard deviation or the standard error of the mean.
        \item It is OK to report 1-sigma error bars, but one should state it. The authors should preferably report a 2-sigma error bar than state that they have a 96\% CI, if the hypothesis of Normality of errors is not verified.
        \item For asymmetric distributions, the authors should be careful not to show in tables or figures symmetric error bars that would yield results that are out of range (e.g., negative error rates).
        \item If error bars are reported in tables or plots, the authors should explain in the text how they were calculated and reference the corresponding figures or tables in the text.
    \end{itemize}

\item {\bf Experiments compute resources}
    \item[] Question: For each experiment, does the paper provide sufficient information on the computer resources (type of compute workers, memory, time of execution) needed to reproduce the experiments?
    \item[] Answer: \answerYes{} 
    \item[] Justification: We state this detailed information of computer resources in Sec.~\ref{sec:id} of the Mainbody .
    \item[] Guidelines:
    \begin{itemize}
        \item The answer \answerNA{} means that the paper does not include experiments.
        \item The paper should indicate the type of compute workers CPU or GPU, internal cluster, or cloud provider, including relevant memory and storage.
        \item The paper should provide the amount of compute required for each of the individual experimental runs as well as estimate the total compute. 
        \item The paper should disclose whether the full research project required more compute than the experiments reported in the paper (e.g., preliminary or failed experiments that didn't make it into the paper). 
    \end{itemize}
    
\item {\bf Code of ethics}
    \item[] Question: Does the research conducted in the paper conform, in every respect, with the NeurIPS Code of Ethics \url{https://neurips.cc/public/EthicsGuidelines}?
    \item[] Answer: \answerYes{} 
    \item[] Justification: We confirm that the research involved in the article complies with the NeurIPS Code of Ethics in all respects.
    \item[] Guidelines:
    \begin{itemize}
        \item The answer \answerNA{} means that the authors have not reviewed the NeurIPS Code of Ethics.
        \item If the authors answer \answerNo, they should explain the special circumstances that require a deviation from the Code of Ethics.
        \item The authors should make sure to preserve anonymity (e.g., if there is a special consideration due to laws or regulations in their jurisdiction).
    \end{itemize}

\item {\bf Broader impacts}
    \item[] Question: Does the paper discuss both potential positive societal impacts and negative societal impacts of the work performed?
    \item[] Answer: \answerYes{} 
    \item[] Justification: The paper have conducted a discussion of broader impacts in the Appendix.
    \item[] Guidelines:
    \begin{itemize}
        \item The answer \answerNA{} means that there is no societal impact of the work performed.
        \item If the authors answer \answerNA{} or \answerNo, they should explain why their work has no societal impact or why the paper does not address societal impact.
        \item Examples of negative societal impacts include potential malicious or unintended uses (e.g., disinformation, generating fake profiles, surveillance), fairness considerations (e.g., deployment of technologies that could make decisions that unfairly impact specific groups), privacy considerations, and security considerations.
        \item The conference expects that many papers will be foundational research and not tied to particular applications, let alone deployments. However, if there is a direct path to any negative applications, the authors should point it out. For example, it is legitimate to point out that an improvement in the quality of generative models could be used to generate Deepfakes for disinformation. On the other hand, it is not needed to point out that a generic algorithm for optimizing neural networks could enable people to train models that generate Deepfakes faster.
        \item The authors should consider possible harms that could arise when the technology is being used as intended and functioning correctly, harms that could arise when the technology is being used as intended but gives incorrect results, and harms following from (intentional or unintentional) misuse of the technology.
        \item If there are negative societal impacts, the authors could also discuss possible mitigation strategies (e.g., gated release of models, providing defenses in addition to attacks, mechanisms for monitoring misuse, mechanisms to monitor how a system learns from feedback over time, improving the efficiency and accessibility of ML).
    \end{itemize}
    
\item {\bf Safeguards}
    \item[] Question: Does the paper describe safeguards that have been put in place for responsible release of data or models that have a high risk for misuse (e.g., pre-trained language models, image generators, or scraped datasets)?
    \item[] Answer: \answerNA{} 
    \item[] Justification: We utilize the publicly available personalized image generation dataset \cite{ruiz2023dreambooth,wang2024instantstyle1,gal2022image,pengdreambench++,gao2024styleshot,huang2025t2icompbench++} and the pretrained generation models Stable Diffusion XL, there is no relevant description we will set up safeguards when we release the code of \textbf{Dual-FDM}.
    \item[] Guidelines:
    \begin{itemize}
        \item The answer \answerNA{} means that the paper poses no such risks.
        \item Released models that have a high risk for misuse or dual-use should be released with necessary safeguards to allow for controlled use of the model, for example by requiring that users adhere to usage guidelines or restrictions to access the model or implementing safety filters. 
        \item Datasets that have been scraped from the Internet could pose safety risks. The authors should describe how they avoided releasing unsafe images.
        \item We recognize that providing effective safeguards is challenging, and many papers do not require this, but we encourage authors to take this into account and make a best faith effort.
    \end{itemize}

\item {\bf Licenses for existing assets}
    \item[] Question: Are the creators or original owners of assets (e.g., code, data, models), used in the paper, properly credited and are the license and terms of use explicitly mentioned and properly respected?
    \item[] Answer: \answerYes{} 
    \item[] Justification: We utilize the publicly available personalized image generation dataset \cite{ruiz2023dreambooth,wang2024instantstyle1,gal2022image,pengdreambench++,gao2024styleshot,huang2025t2icompbench++} and the pretrained generation models Stable Diffusion XL, both of them state the dataset and model license.
    \item[] Guidelines:
    \begin{itemize}
        \item The answer \answerNA{} means that the paper does not use existing assets.
        \item The authors should cite the original paper that produced the code package or dataset.
        \item The authors should state which version of the asset is used and, if possible, include a URL.
        \item The name of the license (e.g., CC-BY 4.0) should be included for each asset.
        \item For scraped data from a particular source (e.g., website), the copyright and terms of service of that source should be provided.
        \item If assets are released, the license, copyright information, and terms of use in the package should be provided. For popular datasets, \url{paperswithcode.com/datasets} has curated licenses for some datasets. Their licensing guide can help determine the license of a dataset.
        \item For existing datasets that are re-packaged, both the original license and the license of the derived asset (if it has changed) should be provided.
        \item If this information is not available online, the authors are encouraged to reach out to the asset's creators.
    \end{itemize}

\item {\bf New assets}
    \item[] Question: Are new assets introduced in the paper well documented and is the documentation provided alongside the assets?
    \item[] Answer: \answerNA{} 
    \item[] Justification: We do not introduce new assets in the paper.
    \item[] Guidelines:
    \begin{itemize}
        \item The answer \answerNA{} means that the paper does not release new assets.
        \item Researchers should communicate the details of the dataset\slash code\slash model as part of their submissions via structured templates. This includes details about training, license, limitations, etc. 
        \item The paper should discuss whether and how consent was obtained from people whose asset is used.
        \item At submission time, remember to anonymize your assets (if applicable). You can either create an anonymized URL or include an anonymized zip file.
    \end{itemize}

\item {\bf Crowdsourcing and research with human subjects}
    \item[] Question: For crowdsourcing experiments and research with human subjects, does the paper include the full text of instructions given to participants and screenshots, if applicable, as well as details about compensation (if any)? 
    \item[] Answer: \answerNA{} 
    \item[] Justification: The paper does not involve crowdsourcing nor research with human subjects.
    \item[] Guidelines:
    \begin{itemize}
        \item The answer \answerNA{} means that the paper does not involve crowdsourcing nor research with human subjects.
        \item Including this information in the supplemental material is fine, but if the main contribution of the paper involves human subjects, then as much detail as possible should be included in the main paper. 
        \item According to the NeurIPS Code of Ethics, workers involved in data collection, curation, or other labor should be paid at least the minimum wage in the country of the data collector. 
    \end{itemize}

\item {\bf Institutional review board (IRB) approvals or equivalent for research with human subjects}
    \item[] Question: Does the paper describe potential risks incurred by study participants, whether such risks were disclosed to the subjects, and whether Institutional Review Board (IRB) approvals (or an equivalent approval/review based on the requirements of your country or institution) were obtained?
    \item[] Answer: \answerNA{} 
    \item[] Justification: The paper does not involve crowdsourcing nor research with human subjects.
    \item[] Guidelines:
    \begin{itemize}
        \item The answer \answerNA{} means that the paper does not involve crowdsourcing nor research with human subjects.
        \item Depending on the country in which research is conducted, IRB approval (or equivalent) may be required for any human subjects research. If you obtained IRB approval, you should clearly state this in the paper. 
        \item We recognize that the procedures for this may vary significantly between institutions and locations, and we expect authors to adhere to the NeurIPS Code of Ethics and the guidelines for their institution. 
        \item For initial submissions, do not include any information that would break anonymity (if applicable), such as the institution conducting the review.
    \end{itemize}

\item {\bf Declaration of LLM usage}
    \item[] Question: Does the paper describe the usage of LLMs if it is an important, original, or non-standard component of the core methods in this research? Note that if the LLM is used only for writing, editing, or formatting purposes and does \emph{not} impact the core methodology, scientific rigor, or originality of the research, declaration is not required.
    \item[] Answer: \answerNA{} 
    \item[] Justification: The paper does not involve LLM as any important, original, or non-standard components.
    \item[] Guidelines:
    \begin{itemize}
        \item The answer \answerNA{} means that the core method development in this research does not involve LLMs as any important, original, or non-standard components.
        \item Please refer to our LLM policy in the NeurIPS handbook for what should or should not be described.
    \end{itemize}

\end{enumerate}